\documentclass[fleqn,10pt]{wlscirep}
\usepackage[utf8]{inputenc}
\usepackage[T1]{fontenc}

\usepackage{float, graphicx, subcaption}
\usepackage{changes}
\usepackage{array}
\newcommand{\PreserveBackslash}[1]{\let\temp=\\#1\let\\=\temp}
\newcolumntype{C}[1]{>{\PreserveBackslash\centering}p{#1}}
\newcolumntype{R}[1]{>{\PreserveBackslash\raggedleft}p{#1}}
\newcolumntype{L}[1]{>{\PreserveBackslash\raggedright}p{#1}}
\usepackage{multirow, diagbox, makecell}
\usepackage{hyperref}
\makeatletter
\newcommand\setcurrentname[1]{\def\@currentlabelname{#1}}
\makeatother

\usepackage{stackengine}

\usepackage{algorithm2e}
\SetKwComment{Comment}{/* }{ */}
\SetKw{Continue}{continue}
\usepackage{amsmath, siunitx}
\usepackage{nameref}
\usepackage{subfiles}

\DeclareFontFamily{U}{mathx}{\hyphenchar\font45}
\DeclareFontShape{U}{mathx}{m}{n}{
      <5> <6> <7> <8> <9> <10>
      <10.95> <12> <14.4> <17.28> <20.74> <24.88>
      mathx10
      }{}
\DeclareSymbolFont{mathx}{U}{mathx}{m}{n}
\DeclareMathSymbol{\bigtimes}{1}{mathx}{"91}

\newcommand{\supplementarysection}{%
  \setcounter{figure}{0}% Reset figure counter
  \setcounter{table}{0}% Reset table counter
  \let\oldthefigure\thefigure% Capture figure numbering scheme
  \let\oldthetable\thetable% Capture table numbering scheme
  \renewcommand{\thefigure}{S\oldthefigure}% Prefix figure number with S
  \renewcommand{\thetable}{S\oldthetable}% Prefix table number with S
  \section*{Supplementary Information}% Set supplementary section
  \let\oldsection\section% Copy \chapter into \oldchapter
  \renewcommand{\section}{% Update \chapter
    \let\thefigure\oldthefigure% Copy \thefigure into \oldthefigure
    \let\thetable\oldthetable% Copy \thetable into \oldthetable
    \let\section\oldsection% Restore original \chapter
    \oldsection% Call original \chapter
  }
}

\title{Topological properties and organizing principles of semantic networks}

\author[1,+]{Gabriel Budel}
\author[1,+]{Ying Jin}
\author[1]{Piet Van Mieghem}
\author[1,*]{Maksim Kitsak}
\affil[1]{Faculty of Electrical Engineering, Mathematics and Computer Science, Delft University of Technology, 2628 CD, Delft, Netherlands}

\affil[*]{M.A.Kitsak@tudelft.nl}

\affil[+]{these authors contributed equally to this work}


\begin{abstract}
Interpreting natural language is an increasingly important task in computer algorithms due to the growing availability of unstructured textual data. Natural Language Processing (NLP) applications rely on semantic networks for structured knowledge representation. The fundamental properties of semantic networks must be taken into account when designing NLP algorithms, yet they remain to be structurally investigated. We study the properties of semantic networks from ConceptNet, defined by 7 semantic relations from 11 different languages. We find that semantic networks have universal basic properties: they are sparse, highly clustered, and many exhibit power-law degree distributions. Our findings show that the majority of the considered networks are scale-free. Some networks exhibit language-specific properties determined by grammatical rules, for example networks from highly inflected languages, such as e.g. Latin, German, French and Spanish, show peaks in the degree distribution that deviate from a power law. We find that depending on the semantic relation type and the language, the link formation in semantic networks is guided by different principles. In some networks the connections are similarity-based, while in others the connections are more complementarity-based. Finally, we demonstrate how knowledge of similarity and complementarity in semantic networks can improve NLP algorithms in missing link inference.
\end{abstract}

\begin{document}

% Mandatory commands template
\flushbottom
\maketitle
\thispagestyle{empty}

%%% INTRODUCTION %%%%%%%%%%%%%%%%%%%%%%%%%%%%%%%%%%%%%%%%%%%%%
\section*{Introduction}\setcurrentname{Introduction}\phantomsection\label{sec:intro}

%% big picture/background 
Due to the explosive increase in the availability of digital content, the demand for computers to efficiently handle textual data has never been greater. Large amounts of data and improved computing power have enabled a vast amount of research on Natural Language Processing (NLP). The goal of NLP is to allow computer programs to interpret and process unstructured text. In computers, text is represented as a string, while in reality, human language is much richer than just a string. People relate text to various concepts based on previously acquired knowledge. To effectively interpret the meaning of a text, a computer must have access to a considerable knowledge base related to the domain of the topic~\cite{cambria2014jumping}.

\begin{figure}[h]
    \centering
    \includegraphics[width=0.4\textwidth]{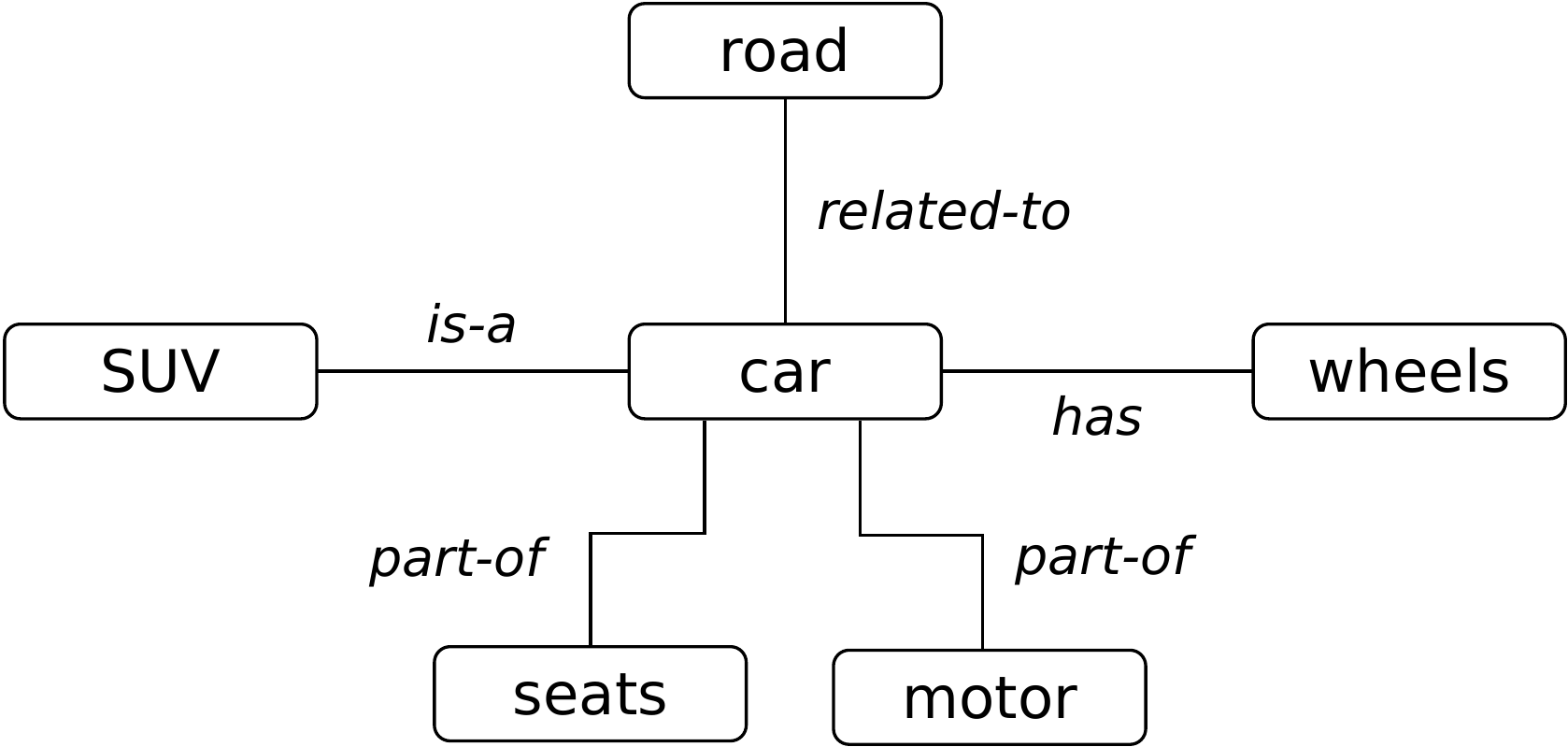}
    \caption{Toy example of a semantic network with six concepts and five semantic relations of four different types.}
    \label{fig:sn}
\end{figure}

%% history & definition
Semantic networks can represent human knowledge in computers, as first proposed by Quillian in the 1960s~\cite{quillian1967word, quillian1969teachable}. `Semantic' means `relating to meaning in language or logic' and a semantic network is a graph representation of structured knowledge. Such networks are composed of nodes, which represent concepts (\textit{e.g.}, words or phrases), and links, which represent semantic relations between the nodes~\cite{01sowa2012semantic, de2022systematic}. The links are tuples of the format \textit{(source, semantic relation, destination)} that encode knowledge. For example, the information that a car has wheels is represented as \textit{(car, has, wheels)}. Figure~\ref{fig:sn} shows a toy example of a semantic network as the subgraph with the neighborhood around the node \textit{car}. 

The past two decades have witnessed a rise in the importance of NLP applications~\cite{sowa2014principles, peters2003using, salem2008ontology}. For instance, Google introduced Google Knowledge Graph to enhance their search engine results~\cite{singhal_2012}. A knowledge graph is a specific type of semantic network, in which the relation types are more explicit~\cite{popping2003knowledge,fensel2020introduction}. Voice assistants and digital intelligence services, such as Apple Siri~\cite{kepuska2018next} and IBM Watson~\cite{high2012era}, use semantic networks as a knowledge base for retrieving information~\cite{piskorski2013information, shi2017data}. As a result, machines can process information in raw text, comprehend unstructured user input, and achieve the goal of communicating with users, all up to a certain extent. Recently, OpenAI made a great leap forward in user-computer interaction with InstructGPT, better known to the general public as ChatGPT~\cite{ouyang2022training}. 

Language is a complex system with diverse grammatical rules. To grasp the meaning of a sentence, humans leverage their natural understanding of language and concepts in contexts. Language is still poorly understood from a computational perspective and, hence, it is difficult for computers to utilize similar strategies. Namely, machines operate under unambiguous instructions that are strictly predefined and structured by humans. Though we can argue that human languages are structured by grammar, these grammatical rules often prove to be ambiguous~\cite{resnik1999semantic}. After all, in computer languages, there are no synonyms, namesakes, or tones that can lead to misinterpretation~\cite{harris_2018}. Computers rely on external tools to enable the processing of the structure and meaning of texts. 	

%% What we do
In this paper, we conduct systematic analyses of the topological properties of semantic networks.
%% Why we do: motivation/reason for this thesis
Our work is motivated by the following purposes:
\begin{itemize}
\item {Understand fundamental formation principles of semantic networks.}\\
In many social networks connections between nodes are driven by similarity~\cite{mcpherson2001birds, kossinets2009origins, schaefer2010fundamental, snijders2011statistical}. The more similar two nodes are in terms of common neighbors, the more likely they are connected. Thanks to the intensive study of similarity-based networks, many successful tools of data analysis and machine learning were developed, such as link prediction~\cite{hasan2011survey} and community detection~\cite{zarandi2018community}. These tools may not work well for semantic networks, because words in a sentence do not necessarily pair together because of similarity. Sometimes, two words are used in conjunction because they have complementary features. Therefore, we study the principles that drive the formation of links in semantic networks.  

\item {Document language-specific features.}\\
Languages vary greatly between cultures and across time~\cite{evans2009myth}. Two languages that originate from two different language families can differ in many types of features since they are based on different rules. It is natural to conjecture that there exist diverse structures in semantic networks for different languages. 

\item {Better inform NLP methods.}\\
Although there have been numerous real-world NLP applications across various domains, existing NLP technologies still have limitations~\cite{khurana2022natural}. For example, processing texts from a language where single words or phrases can convey more than one meaning is difficult for computers~\cite{alfawareh2011resolving, jusoh2018study}. Existing, successful algorithms built on top of semantic networks are usually domain-specific, and designing algorithms for broader applications remains an open problem. To design better language models that can handle challenges such as language ambiguity, we first need to gain a better understanding of the topological properties of semantic networks. 

\end{itemize}

%% research gap
Previous studies on semantic networks focused on a few basic properties and relied on multiple datasets with mixed semantic relations, which we discuss in detail in the `\nameref{sec:rel_work}' section. Therefore, it is difficult to compare the results within one study and between two different studies. To our knowledge, there has been no systematic and comprehensive analysis of the topological properties of semantic networks at the semantic relation level.

%% goal
To sum up, the main objective of this paper is to understand the structure of semantic networks. Specifically, we first study the general topological properties of semantic networks from a single language with distinct semantic relation types. Second, we compare semantic networks with the same relation type between different languages to find language-specific patterns. In addition, we investigate the roles of similarity and complementarity in the link formation principles in semantic networks.

The main contributions of this paper include:
\begin{enumerate}
\item We study the topological properties of seven English semantic networks, each network defined by a different semantic relation (\textit{e.g.}, `Is-A' and `Has-A'). We show that all networks possess high sparsity and many possess a power-law degree distribution. In addition, we find that most networks have a high average clustering coefficient, while some networks show the opposite. 

\item We extend the study of the topological properties of semantic networks to ten other languages. We find non-trivial structural patterns in networks from languages that have many grammatical inflections. Due to the natural structure of grammar in these languages, words have many distinct inflected forms, which leads to peaks in the density of the degree distribution and results in deviations from a power law. We find this feature not only in inflecting languages but also in Finnish, which is classified as agglutinating.

\item We study the organizing principles of 50 semantic networks defined by different semantic relations in different languages. We quantify the structural similarity and complementarity of semantic networks by counting the relative densities of triangles and quadrangles in the graphs, following a recent work by Talaga and Nowak~\cite{talaga2022}. Hereby, we show to what extent these networks are similarity- or complementarity-based. We find that the connection principles in semantic networks are mostly related to the type of semantic relation, not the language origin. 
\end{enumerate}

This paper is organized in the following manner: 
In the `\nameref{sec:rel_work}' section, we provide a brief overview of the previous work on the properties of semantic networks. In the `\nameref{sec:gen_properties}' section, we study the general topological properties of seven English semantic networks. In the `\nameref{sec:diff_languages}' section, we compare the properties of semantic networks between 11 different languages. The section `\nameref{sec:simil_compl}' deals with the fundamental connection principles in semantic networks. We measure and compare the structural similarity and complementarity in the networks in this study and we discuss the patterns that arise. Finally, we summarize our conclusions and findings and give recommendations for future research in the `\nameref{sec:discussion}' section.

%%% RELATED WORK %%%%%%%%%%%%%%%%%%%%%%%%%%%%%%%%%%%%%%%%%%%%%
\section*{Related work}\setcurrentname{Related work}\phantomsection\label{sec:rel_work}
Due to the growing interest in semantic networks, related studies were carried out in a wide range of different fields. Based on our scope, we focus on two main aspects in each work: the topological properties that were analyzed in the study and the dataset that was used in the analysis (i) and the universal and language-specific patterns which were found and discussed (ii).

The majority of semantic networks literature is centered around three link types: co-occurrence, association and semantic relation. In a co-occurrence network, sets of words that co-occur in a phrase, sentence or text form a link. For association networks, participants in a cognitive-linguistic experiment are given a word and asked to give the first word that they think of. There are several association datasets, one example is the University of South Florida Free Association Norms~\cite{nelson1999university}. Semantic relations are relations defined by professionals like lexicographers, typical examples are synonym, antonym, hypernym and homonymy. The specific instances of the semantic relations are also defined by the lexicographers or extracted computationally from text corpora.

In 2001, Ferrer-i-Cancho and Sole~\cite{cancho2001small} studied undirected co-occurrence graphs constructed from the British National Corpus dataset~\cite{bnc2007british}. They measured the average distance between two words and observed the small-world property, which was found in many real-world networks~\cite{watts1998collective}. Motter \textit{et al.}~\cite{motter2002topology} analyzed an undirected conceptual network constructed from an English Thesaurus dictionary~\cite{ward2015moby}. They focused on three properties: sparsity (small average degree), average shortest path length and clustering. That same year, Sigman and Cecchi~\cite{sigman2002global} studied undirected lexical networks extracted from the noun subset of WordNet~\cite{miller1995wordnet}, where the nodes are sets of noun synonyms. They grouped networks by three semantic relations: antonymy, hypernymy and meronymy. A detailed analysis of characteristic length (the median minimal distance between pairs of nodes), degree distributions and clustering of these networks were provided. Semantic networks were also found to possess the small-world property of sparse connectivity, short average path length, and strong local clustering~\cite{motter2002topology, sigman2002global}.

Later, Steyvers and Tenenbaum~\cite{steyvers2005large} performed statistical analysis of 3 kinds of semantic networks: word associations~\cite{nelson1999university}, WordNet and Roget’s Thesaurus~\cite{roget1911roget}. Apart from the above-mentioned network properties, they also considered network connectedness and diameter. They pointed out that the small-world property may originate from the scale-free organization of the network, which exists in a variety of real-world systems~\cite{barabasi1999emergence, strogatz2001exploring,VanMieghem2014}.

As for patterns across different languages, Ferrer-i-Cancho \textit{et al.}~\cite{cancho2004patterns} built syntactic dependency networks from corpora (collections of sentences) for three European languages: Czech, German and Romanian.
They showed that networks from different languages have many non-trivial topological properties in common, such as the small-world property, a power-law degree distribution and disassortative mixing~\cite{noldus2015assortativity}. 

% Research gap
Existing studies have identified some general network properties in semantic networks such as the small-world property and power-law degree distributions. However, the datasets used in these studies are often different, sometimes even within the same study, rendering direct comparison of results difficult. Some used associative networks generated from experiments and some studied thesauri that were manually created by linguists. In addition, most of the research performed consists of coarse-grained statistical analyses. Specifically, different semantic relations were sometimes treated as identical and the subset of included nodes was often limited (\textit{e.g.}, only words and no phrases or only nouns). Further, there are only very few studies on semantic networks from languages other than English.

% goal
Therefore, our analyses focus on semantic networks with different semantic relations (link types) from a single dataset. We consider networks defined by a specific link type, make these networks \textit{undirected} and \textit{unweighted} and compare the structural properties between networks with different link types. In addition, we apply similar analyses to semantic networks with the same link type across different languages. Furthermore, we investigate the roles that similarity and complementarity play in the formation of links in semantic networks.

%%% GENERAL PROPERTIES %%%%%%%%%%%%%%%%%%%%%%%%%%%%%%%%%%%%%%%%%%%%%

\section*{General properties of semantic networks}\setcurrentname{General properties of semantic networks}\phantomsection\label{sec:gen_properties}

To gain an understanding of the structure of semantic networks, we first study their general topological properties. We introduce the main characteristics of the dataset that we use throughout this study, ConceptNet~\cite{ConceptNet}, in the section `\nameref{sec:SI_data}' in SI. Next, we list the semantic relations that define the networks in this study in Supplementary Table~\ref{tab:defrels}. The overview of the semantic networks is given in Supplementary Table~\ref{tab:fullstats}.
In this section, we compute various topological properties of these networks related to connectedness, degree, assortative mixing and clustering. We summarize the overall descriptive statistics of the semantic networks in Supplementary Table~\ref{tab:statslcc}.

\subsection*{Connectedness}
We measure the connectedness of a network by the size of the largest connected component and the size distribution of all connected components. The complete component size distributions of the English semantic networks are shown in Supplementary Figure~\ref{fig:ccsize}. Supplementary Table~\ref{tab:lcc} lists the sizes of the largest connected components (LCCs) in absolute numbers as well as relative to the network size. The same statistics are computed after degree-preserving random rewiring of the links for comparison\cite{noldus2015assortativity}. The purpose of random rewiring is to estimate the value of a graph metric that could be expected by chance, solely based on the node degrees (see SI for details on the rewiring process). 

Based on the percentages of nodes in the LCC, all seven semantic networks are not fully connected. The networks `Is-A', `Related-To' and `Union' are almost fully connected given that their LCCs contain over 90\% of nodes. Networks `Has-A', `Part-Of', `Antonym' and `Synonym' are largely disconnected, with the percentages of nodes in their LCCs ranging from 22\% to 65\%. Most of the rewired networks are more connected than the corresponding original networks, especially networks `Antonym' and `Synonym'. In other words, the majority of our semantic networks are less connected than what could be expected by chance. For networks `Related-To' and `Union', the percentage of nodes in the LCC remains almost unchanged, while the `Is-A' network is more connected than expected.

\subsection*{Degree distribution}
\begin{figure}[h]
      \centering
      \includegraphics[width=\textwidth]{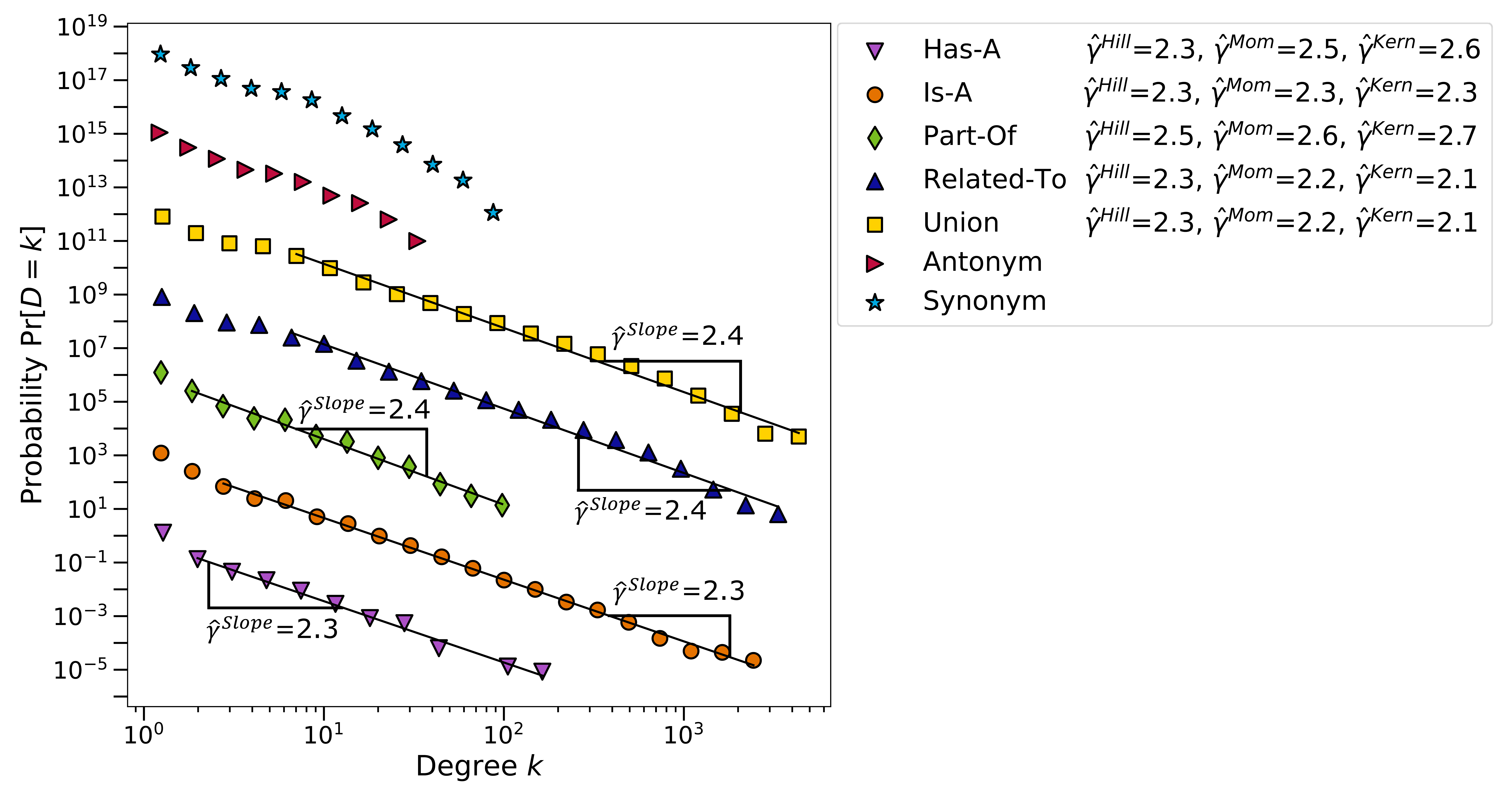}
    \caption{Degree distribution densities $\Pr[D=k]$ of the LCCs of the seven English semantic networks. The data is scaled by powers of 1000 to better visualize the power law in each density. The corresponding estimated power-law exponents $\hat{\gamma}$ are shown if there is a power law, $\Pr[D=k] \approx \ell(k) k^{-\gamma}$. The degree sequences of the networks `Antonym' and `Synonym' were estimated to be \textit{hardly power-law} because at least one of the $\hat{\gamma} > 5$. The data are logarithmically binned to suppress noise in the tails of the distributions, see the section `\nameref{sec:power_laws}' in SI for details on how the power-law densities are processed and the power-law exponent estimation procedures.}
    \label{fig:deg1}
\end{figure}

Figure~\ref{fig:deg1} shows that the densities $\Pr[D=k]$ of the degree distributions of our seven English semantic networks all appear to approximately follow power laws in the tail visually. A more rigorous framework for assessing power laws was proposed by Voitalov~\textit{et~al.}~\cite{voitalov2019scale}, who consider networks to have a power-law degree distribution if $\Pr[D=k] = \ell(k) k^{-\gamma}$ for a slowly varying function $\ell(k)$, see the section `\nameref{sec:consistent_powerlaw}' in SI. Figure~\ref{fig:deg1} includes the estimates $\hat{\gamma}$ based on the slopes of the densities $\Pr[D=k]$ on a log-log scale, along with the three consistent estimators from the framework of Voitalov~\textit{et~al.}~\cite{voitalov2019scale, voitalov2019code}. According to these estimators, the degree sequences of 5 out of the 7 networks are power-law. The degree sequences of the `Synonym' and `Antonym' networks are \textit{hardly power-law} because at least one of the $\hat{\gamma} > 5$ and therefore the estimated exponents are not listed.

For most networks, the estimated exponent $\hat{\gamma}$ lies between 2 and 3. Therefore, most semantic networks are scale-free, except for the `Synonym' and `Antonym' networks. In the literature, semantic networks were also found to be highly heterogeneous~\cite{steyvers2005large, borge2010semantic}. Moreover, the word frequencies in several modern languages were found to follow power laws~\cite{petersen2012languages}. In the section `\nameref{sec:diff_languages}', we will see that the `Synonym' and `Antonym' networks in most considered languages are hardly power-law or not power-law networks. 

The heterogeneity in the degree distribution seems natural for networks such as the `Is-A' network: there are many specific or unique words with a small degree that connect to only a few other words, while there are also a few general words that connect to almost anything, resulting in a large degree. Examples of general words with a large degree are `plant' and `person', while specific words like `neotectonic' and `cofinance' have a small degree. Our results show that many semantic networks have power-law degree distributions, like many other types of real-world networks~\cite{adamic2000power, jeong2000large, faloutsos1999power}. 

\subsection*{Degree assortativity}
\begin{figure}[h]
		\includegraphics[width=1\textwidth]{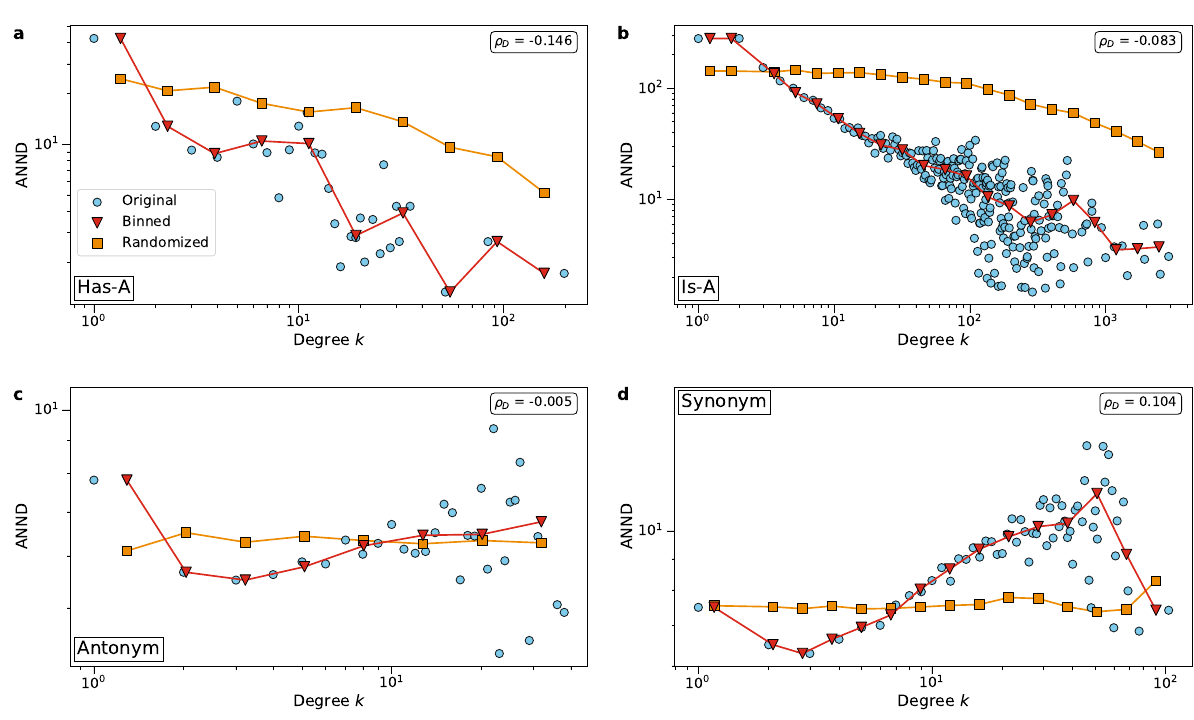}
	\caption{Average nearest neighbor degree (ANND) as a function of degree $k$ and degree correlation coefficient $\rho_D$ of four English semantic networks. (a) Network \textit{`Has-A'}, (b) Network \textit{`Is-A'}, (c) Network \textit{`Antonym'}, (d) Network \textit{`Synonym'}. See SI for the results of all seven networks. Data points in circles are the average ANND of nodes with degree $k$ in a network, triangles represent the data after logarithmic binning, and squares are the average ANND of nodes with degree $k$ in the rewired network. Logarithmic binning is to better visualize the data.}
	\label{fig:annd}
	\end{figure}

\begin{figure}[H]
    \includegraphics[width=1\textwidth]{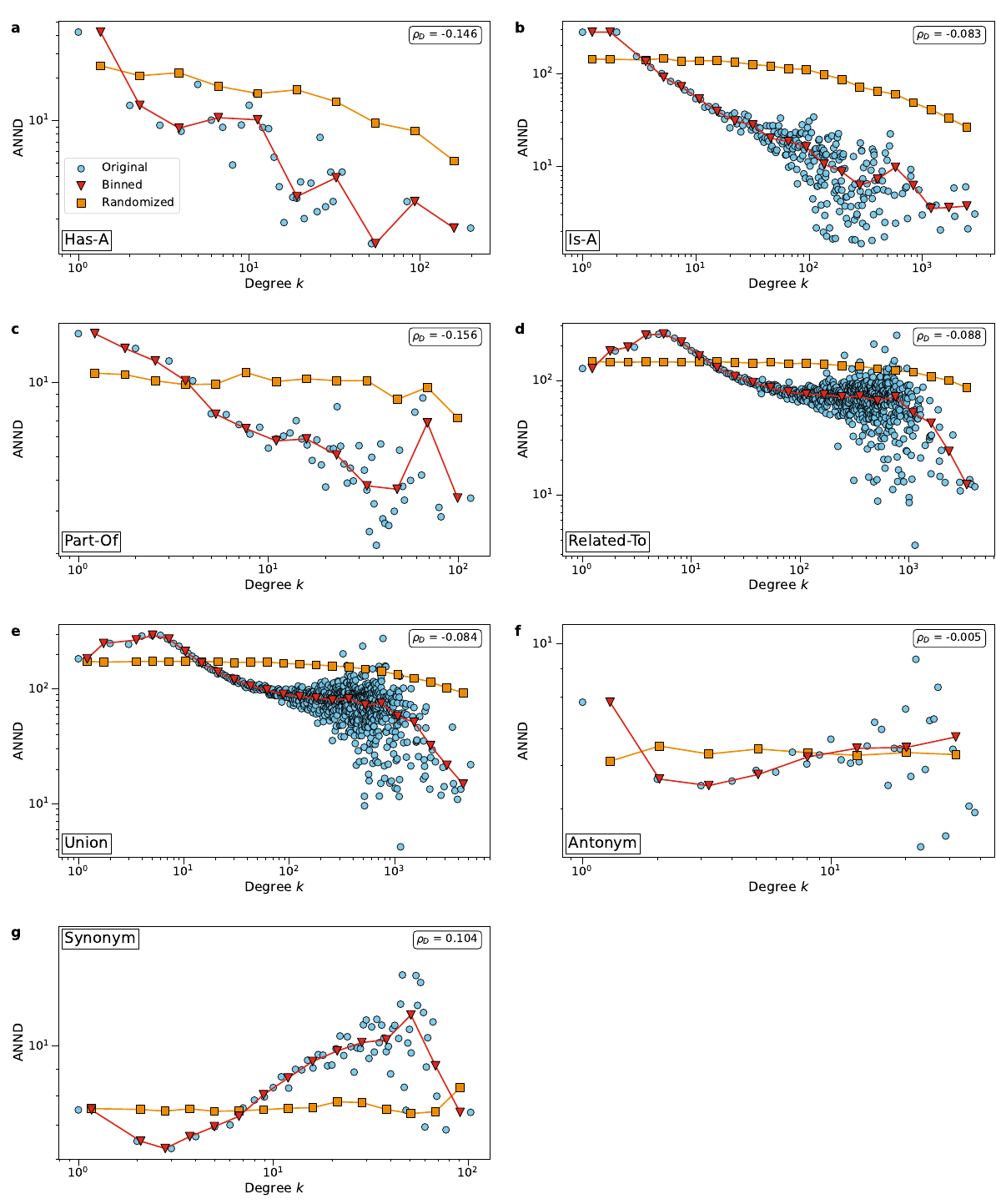}
    \caption{Average nearest neighbor degree (ANND) as a function of the degree $k$ and degree correlation coefficient $\rho_D$ of the seven English semantic networks. (a) Network \textit{`Has-A'}, (b) Network \textit{`Is-A'}, (c) Network \textit{`Part-Of'}, (d) Network \textit{`Related-To'}, (e) Network \textit{`Union'}, (f) Network \textit{`Antonym'}, (g) Network \textit{`Synonym'}. The circle data points are the original average ANND of nodes with degree $k$ in a network, triangles represent the data after logarithmic binning, and squares are the average ANND of nodes with degree $k$ in the randomized network.}
    \label{fig:annd1}
\end{figure}

\phantomsection\label{sec:assort}
A number of measures have been established to quantify degree assortativity, such as the degree correlation coefficient $\rho_D$ and the Average Nearest Neighbor Degree (ANND) \cite{noldus2015assortativity}. Figure~\ref{fig:annd} shows the average nearest neighbor degree as a function of the degree $k$ for four selected networks and their values after random rewiring as well as the degree correlation coefficient~$\rho_D$. Refer to Supplementary Fig.~\ref{fig:annd1} for the ANND plots of all networks. The randomized networks with preserved degree distribution have no degree-degree correlation. As a result, the function ANND does not vary with $k$. The randomized networks serve as a reference for the expected ANND values when the links are distributed at random.

We find that most semantic networks are disassortative as ANND is a decreasing function of the degree $k$ and the degree correlation coefficient $\rho_D$ is negative. These networks are `Has-A’, `Part-Of’, `Is-A’, `Related-To’ and `Union’. In disassortative networks, nodes with larger degrees (general words) tend to connect to nodes with smaller degrees (less general words). This is not surprising. Indeed, if we use these relations in a sentence, then we often relate specific words to more general words. For example, we say `horse racing is a sport', in which `horse racing' is a very specific phrase while `sport' is more general.

On the other hand, network `Synonym' is assortative as the function ANND increases in the degree $k$. This indicates that large-degree nodes (general words) connect to nodes that have similar degree (words with the same generality). The same applies to network `Antonym'. Although the degree correlation is not very pronounced and reflected by the small correlation coefficient $\rho_D=-0.005$, we still see a slight upward trend in the curve of ANND.

The function ANND of a rewired network is not degree-dependent anymore, shown by the orange curves in Figure~\ref{fig:annd}. The curve is almost flat for `Synonym' and `Related-To'. At the larger degree $k$, the curve may drop slightly, as for large-degree nodes there are not enough nodes of equal degree to connect to.

\subsection*{Clustering coefficient}\label{ch:clustercoef}
In networks such as social networks, the neighbors of a node are likely to be connected as well, a phenomenon which is known as clustering~\cite{watts1998collective,newman2001clustering}. If a person has a group of friends, there is a high chance that these friends also know each other. These networks are characterized by many triangular connections. 

\begin{figure}[h]
		\includegraphics[width=\textwidth]{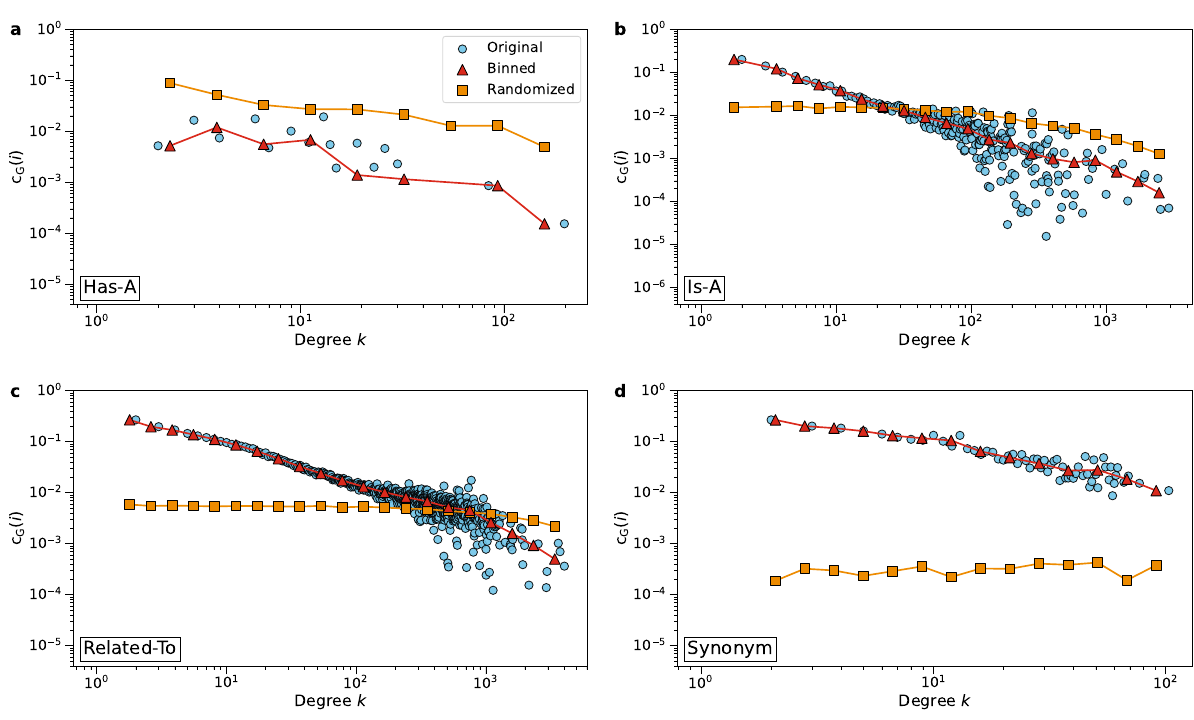}
	\caption{The average clustering coefficient $c_G(i)$ of nodes with degree $d_{i}=k$ of four English semantic networks. (a) Network \textit{`Has-A'}, (b) Network \textit{`Is-A'}, (c) Network \textit{`Related-To'},  (d) Network \textit{`Synonym'}. See the SI for the results of all seven networks. Data points in circles are the original average local clustering coefficients of nodes with degree $d_{i}=k$, triangles represent data after logarithmic binning, and squares show the average clustering coefficients of nodes with degree $d_{i}=k$ (logarithmically binned) in the randomized networks.}
	\label{fig:cg}
	\end{figure}

\begin{figure}[H]
		\includegraphics[width=\textwidth]{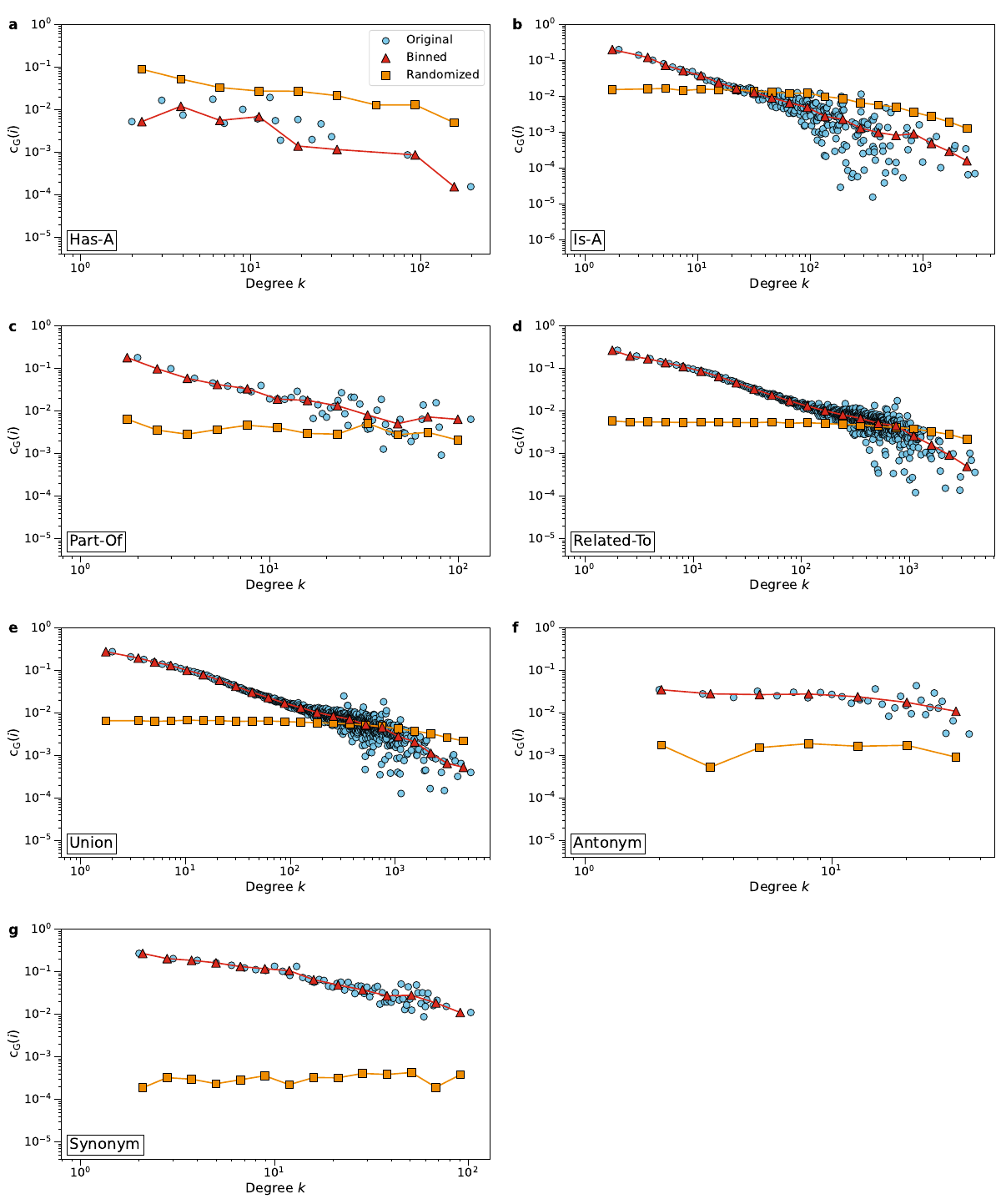}
	\caption{The average clustering coefficient $c_G(i)$ of nodes with degree $d_{i}=k$ of the seven English semantic networks. (a) Network \textit{`Has-A'}, (b) Network \textit{`Is-A'}, (c) Network \textit{`Part-Of'}, (d) Network \textit{`Related-To'}, (e) Network \textit{`Union'}, (f) Network \textit{`Antonym'}, (g) Network \textit{`Synonym'}. The circle data points are the original average local clustering coefficients of nodes with degree $d_{i}=k$, triangles represent data after logarithmic binning, and squares show the average clustering coefficient of nodes with degree $d_{i}=k$ (logarithmically binned) in the randomized networks.}
	\label{fig:cg1}
	\end{figure}

Figure~\ref{fig:cg} shows the average clustering coefficient $c_G(i)$ of nodes with degree $d_{i}=k$ of four English networks. Refer to Supplementary Fig.~\ref{fig:cg1} for the clustering coefficients of all seven networks. All networks have small clustering coefficients in absolute terms, which, in combination with the small average degree $E[D]$, indicates a local tree-like structure. We find that the networks `Part-Of', `Antonym' and `Synonym' have substantially larger clustering coefficients than their rewired counterparts: there are more triangles in these networks than expected by chance. On the other hand, the network `Has-A' has lower clustering coefficients $c_G(i)$ than the randomized network, therefore it seems that the `Has-A’ network is organized in a different way than the other networks. As for the networks `Is-A', `Related-To' and `Union', the clustering coefficients $c_G(i)$ are similar to their corresponding rewired networks.

In summary, we find that English semantic networks have power-law degree distributions and most are scale-free, which coincides with the results in previous studies~\cite{steyvers2005large,borge2010semantic}.
Besides, semantic networks with different link types show different levels of degree assortativity and average local clustering. Most works in the literature have identified high clustering coefficients in semantic networks~\cite{steyvers2005large,borge2010semantic,motter2002topology,sigman2002global}.
This encourages us to further investigate the organizing principles of these semantic networks, which we will discuss explicitly in the `\nameref{sec:simil_compl}' section.

%%% LANGUAGE-SPECIFIC PROPERTIES %%%%%%%%%%%%%%%%%%%%%%%%%%%%%%%%%%%%%%%%%%%%%
\section*{Language-specific properties}\setcurrentname{Language-specific properties}\phantomsection\label{sec:diff_languages}
Up to this point, we have only considered English semantic networks, while there are thousands of other languages in the world besides English. In this section, we consider semantic networks from 10 other languages contained within ConceptNet: French, Italian, German, Spanish, Russian, Portuguese, Dutch, Japanese, Finnish and Chinese. We group the in total 11 languages based on their language families and we again study the topological properties of 7 different semantic relations per language. Finally, we observe peculiarities in the degree distribution densities of the `Related-To' networks in some languages, which we later explain by grammar.

\subsection*{Language classification} \label{langfamily}
In linguistics, languages can be partitioned in multiple different ways. Mainly, there are two kinds of language classifications: genetic and typological. 

The genetic classification assorts languages according to their level of diachronic relatedness, where languages are categorized into the same family if they evolved from the same root language.~\cite{lyons}. An example is the Indo-European family, which includes the Germanic, Balto-Slavic and Italic languages~\cite{eberhard22simons}. 
 
One popular typological classification distinguishes isolating, agglutinating and inflecting languages. It groups languages in accordance with their morphological word formation styles. A morph or morpheme (the Greek word $\mu\omicron\rho\varphi\acute{\eta}$ means `outer shape, appearance' of which the English `form' is derived) is the basic unit of a word, such as a stem or an affix~\cite{haspelmath2020morph}. For instance, the word `undoubtedly' consists of three morphs: `un-', `doubted' and `-ly'. In an isolating language, like Mandarin Chinese, each word contains only a single morph~\cite{lyons}. In contrast, words from an agglutinating language can be divided into morphs with distinctive grammatical categories like tense, person and gender. In an inflecting language, however, there is no exact match between morphs and grammatical categories~\cite{lyons}. A word changes its form based on different grammar rules. Most Indo-European languages belong to the inflecting category. 

Based on these two types of classifications, we have selected 11 languages to cover different language types, Supplementary Table~\ref{tab:lang_class}. Typologically, Chinese is an isolating language, while Japanese and Finnish are agglutinating languages. The rest of the languages under consideration (8 out of 11) belong to the inflecting category. Genetically, French, Italian, Spanish and Portuguese belong to the Italic family, while English, German and Dutch are Germanic. Russian is a Balto-Slavic language, Japanese is Transeurasian, Chinese is Sino-Tibetan and Finnish belongs to the Uralic family. We mainly refer to the typological classification throughout our analyses.

\subsection*{Overview of semantic networks from eleven languages}
For every language, we construct seven undirected semantic networks with the link types `Has-A', `Part-Of', `Is-A', `Related-To', `Union', `Antonym' and `Synonym'. Due to missing data in ConceptNet, only three languages have the `Has-A' relation. For these languages, the `Union' network is the union of three link types: `Part-Of', `Is-A' and `Related-To'. In this section, we provide an overview of the numbers of nodes $N$ and numbers of links $L$ of the semantic networks. Again, we restrict our study to the LCCs of these networks. 

Regarding the numbers of nodes $N$, the networks `Related-To' and `Union' are generally the largest networks in a language, with the French `Union' network being the absolute largest with $N=1,296,622$, as denoted in Supplementary Table~\ref{tab:langs_nodes}. Nevertheless, there are many small networks with size $N<100$, particularly for the `Part-Of' and `Synonym' networks.

Similar to the English semantic networks, we observe that most networks with more than 100 nodes are sparse. All networks have an average degree between 1 and 6, which is small compared to the network size. Consider the Dutch `Is-A' network for example, where a node has about 5 connections on average, which is only 2.45\% of 191 nodes in the whole network. Supplementary Table~\ref{tab:langs_ED} lists the average degree $E[D]$ of all our semantic networks.

\subsection*{Degree distribution}
Many of the semantic networks in the 11 languages have degree distributions that are approximately power laws. We estimate the power-law exponents only for networks with size $N>1000$ because we require a sufficient number of observations to estimate the power-law exponent~$\gamma$. Supplementary Table~\ref{tab:langs_gamma} lists the estimated power-law exponents $\hat{\gamma}$ using the same 4 methods as in Fig.~\ref{fig:deg1} for each semantic network. Refer to the section `\nameref{sec:power_laws}' in SI for details on these estimation procedures.

We find that many networks have power laws in their degree distributions and many of those networks are scale-free ($2 < \hat{\gamma} < 3$). The Chinese `Related-To' network even has a power-law exponent~$\hat{\gamma} <2$. The degree distributions of all `Synonym' and `Antonym' networks are hardly or not power laws, however. The likely reason for this is that nodes in these networks generally have smaller degrees than in other networks. As a result, the slope of the degree distribution is steeper and therefore not classified as a power law. This is not unexpected, as for a given word there are only a certain number of synonyms or antonyms and therefore there are not many nodes with high degrees. Another interesting finding is that the densities of the degree distributions of the `Related-To' and `Union' networks for French, Spanish, Portuguese and Finnish show notable deviations from a straight line in the log-log plot, which we discuss in-depth in the next section.

\subsection*{Language inflection} \label{inflection}
In some languages, the densities of the degree distributions of the `Related-To' and `Union' networks show deviations from a straight line on a log-log scale. An example is the Spanish `Related-To' network in Fig.~\ref{fig:RT2}a, where we observe a peak in the tail of the distribution. To find the cause of the anomaly in the degree distribution, we inspect the words with a degree $k$ located in the peak, referred to as \textit{peak words}, and their neighbors. Supplementary Table~\ref{tab:es_peak} lists a few examples of the peak words, which are almost all verbs and have similar spellings. The links adjacent to these nodes with higher-than-expected degrees might be the result of grammatical inflections of the same root words since Spanish is a highly inflected language. We observe a similar anomaly in the degree distributions of French, Portuguese and Finnish `Related-To' and `Union' networks. In Supplementary Table~\ref{tab:langs_nodes} we saw that the network `Union' is mostly composed of `Related-To' in these four languages, therefore we restrict the analysis to the `Related-To' networks. 

Two common types of language inflection are conjugation, the inflection of verbs, and declension, the inflection of nouns. The past tense of the verb `to sleep' is `slept', an example of conjugation in English. The plural form of the noun `man' is `men', an example of declension. The languages Spanish, Portuguese and French are much richer in conjugations than English, while Finnish is rich in declensions.

\subsubsection*{Part-of-speech tags}
In the ConceptNet dataset, only part of the nodes is part-of-speech (POS) tagged with one of four types: verb, noun, adjective and adverb. For French, Spanish and Portuguese, the percentage of verbs in the peak is larger than in the LCC, while for Finnish the percentage of nouns in the peak is larger than in the LCC, see Supplementary Table~\ref{tab:wordtypes}. Remarkably, 100\% of the Portuguese peak words are verbs. Most neighbors of the peak words are verbs for Spanish (97\%), Portuguese (99\%) and French (87\%), while most neighbors of the peak words are nouns for Finnish (90\%), Supplementary Table~\ref{tab:nb_wordtypes}. This strengthens our belief that the abnormal number of nodes with a certain degree $k$ is related to language inflection in these four languages.

\subsubsection*{Merging of word inflections}
To investigate whether the peaks in the degree distribution densities are indeed related to word inflections, we leverage the `Form-Of' relation type in ConceptNet, which connects two words A and B if A is an inflected form of B, or B is the root word of A~\cite{Conceptnetrelations}. We merge each node and its neighbors from the `Form-Of' network (its inflected forms) into a single node in the `Related-To' network, as depicted in Supplementary Figure~\ref{fig:merge}. Figure~\ref{fig:RT2} shows the degree distribution densities of the `Related-To' networks before and after node merging. The range of the anomalous peak in the density of the degree distribution is highlighted in yellow. In each panel, the number of grammatical variations $m$ coincides with the center of the peak. As seen in Fig.~\ref{fig:RT2}a, the peak completely disappears in the Spanish `Related-To' after node merging, thus the peak is described entirely by connections due to word inflections. We also observe significant reductions in the heights of the peaks for Portuguese and Finnish `Related-To' networks. However, for the French `Related-To' network there is only a slight reduction in height after merging, which we believe is likely due to poor coverage in the French `Form-Of' network with only 17\% of words in the peak. In contrast, the Spanish `Form-Of' network covers 97\% of the Spanish peak words, while for Portuguese and Finnish approximately 50\% of the peak words are covered, Supplementary Table~\ref{tab:FORT2 commonN}.

\begin{figure}[h]
	\includegraphics[width=\textwidth]{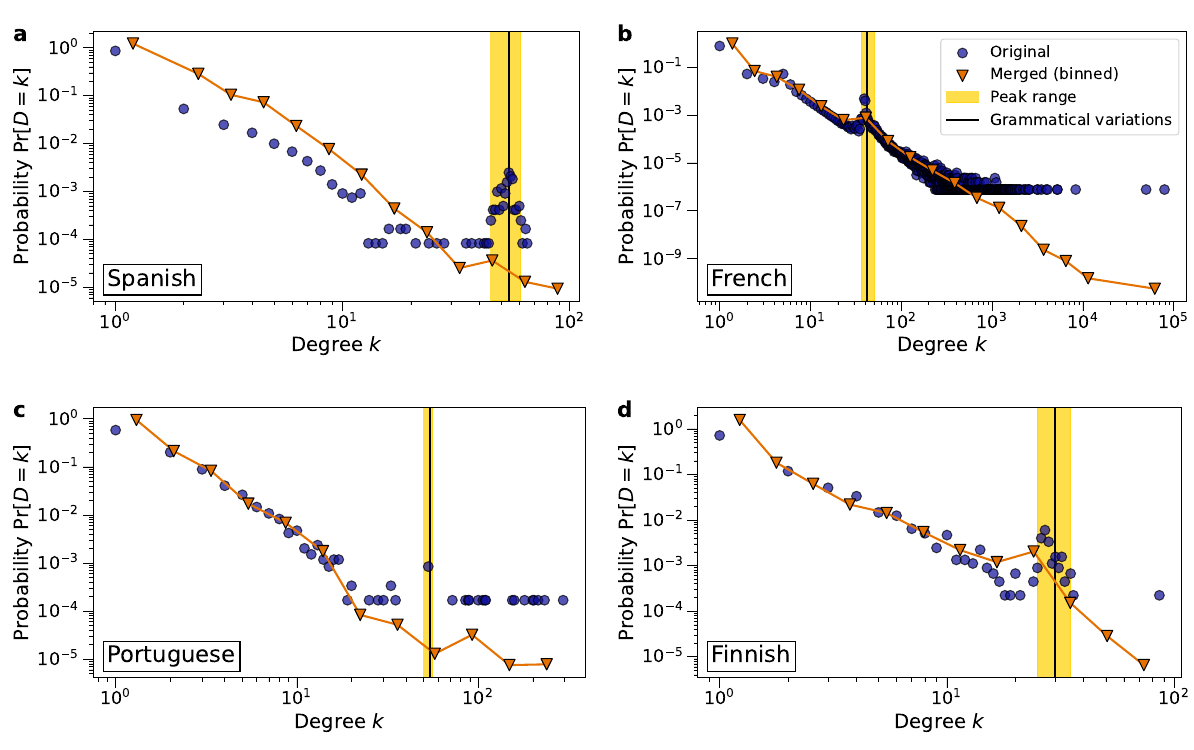}
	\caption{Densities $\Pr[D=k]$ of the degree distributions of the `Related-To' networks before and after node merging of inflected forms in (a) Spanish, (b) French, (c) Portuguese and (d) Finnish. The logarithmically binned densities after node merging are shown in orange. The peaks are highlighted in yellow. The vertical black lines indicate the number of grammatical variations $m$ for the relevant grammatical rule in the respective language. In each panel the number of grammatical variations $m$ coincides with the center of the peak.}
	\label{fig:RT2}
	\end{figure}

\subsubsection*{The number of inflections} \label{grammar}
In a language, the number of distinct conjugations of regular verbs is determined by the number of different pronouns and the number of verb tenses, which are grammatical time references~\cite{comrie1976aspect}. In Spanish, there are 6 pronouns and 9 simple verb tenses, resulting in at most $m = 6 \times 9 = 54$ distinct verb conjugations~\cite{kendris2020501, vare2022conjugation}. Table~\ref{tab:es_grammar} exemplifies these 54 different conjugations for the verb `amar', which means `to love'. There are also irregular verbs that follow different, idiosyncratic grammatical rules, but the majority of the verbs in Spanish are classified as regular, like in most inflecting languages. The number of grammatical variations $m = 54$ coincides with the center of the peak in Figure~\ref{fig:RT2}a. 

\begin{table}[h]
	\centering
	\footnotesize
	\begin{tabular}{lllllll}
	\toprule
	\multirow{2}{*}{\diagbox{Tense}{Pronoun}} & Yo &Tú &Él/Ella/Usted &Nosotros  &Vosotros &Ellos/Ellas/Ustedes\\ 
												& (I) &(You) &(He/She) &(We) & (You) & (They)\\ %\rule{0pt}{12pt}
            \midrule
	Present Indicative & amo &	amas&	ama	&	amamos&	amáis&	aman\\ \hline  %\rule{0pt}{18pt}
	Imperfect Indicative & amaba&	amabas&	amaba&	amábamos&	amabais&	amaban\\ \hline  %\rule{0pt}{18pt}
	Preterite Indicative& amé&	amaste &	amó	& amamos&	amasteis&	amaron\\ \hline  %\rule{0pt}{18pt}
	Future Indicative& amaré&	amarás&	amará&	amaremos&	amaréis&	amarán\\ \hline  %\rule{0pt}{18pt}
	Conditional Indicative & amaría&	amarías&amaría&	amaríamos&	amaríais&	amarían\\ \hline  %\rule{0pt}{18pt}
	Present Subjunctive & ame&	ames&	ame&	amemos&	améis&	amen\\ \hline  %\rule{0pt}{18pt}
	Imperfect Subjunctive 1 & amara&	amaras&	amara&	amáramos&	amarais&	amaran\\ \hline  %\rule{0pt}{18pt}
	Imperfect Subjunctive 2 & amase&	amases&	amase&	amásemos&	amaseis&	amasen\\ \hline  %\rule{0pt}{18pt}
	Future Subjunctive & amare&	amares&	amare&	amáremos&	amareis&	amaren\\
	\bottomrule
	\end{tabular}
	\caption{Verb conjugation table for the Spanish verb `amar' (to love). The 6 pronouns and 9 verb tenses result in a maximum of 54 different conjugated forms.}
	\label{tab:es_grammar}
 \end{table}

Like Spanish, Portuguese has $m = 54$ distinct conjugated verb forms~\cite{nitti2015501}. In French, there are $m = 6 \times 7 = 42$ distinct verb conjugations~\cite{lawless2005everything}. In Finnish, there are in total 15 noun declensions or cases with distinct spelling, each having singular and plural forms, resulting in $m = 30$ different cases of a Finnish noun~\cite{karlsson2017finnish}. Supplementary Table~\ref{tab:kmatch} lists the number of grammatical variations $m$ in French, Spanish, Portuguese and Finnish, along with the minimum and maximum degree $k_{min}$ and $k_{max}$ where the peak starts and ends. By Fig.~\ref{fig:RT2} we confirm that the number of grammatical variations $m$ coincides with or is close to the center of the peak. 

In summary, we observe anomalies in the degree distributions of `Related-To’ networks from the inflecting languages Spanish, French and Portuguese and the agglutinating Finnish. Because of grammatical structures, root words in these languages share many links with their inflected forms, resulting in more nodes with a certain degree than expected. While Finnish is typologically classified as agglutinating, it still has many noun declensions, suggesting that the agglutinating and inflecting language categories are not mutually exclusive.

\section*{Similarity and complementarity in semantic networks}\setcurrentname{Similarity and complementarity in semantic networks}\phantomsection\label{sec:simil_compl}
Although we have identified several universal characteristics of semantic networks, we also observe notable differences in some of their properties. The clustering coefficient in some semantic networks, for instance, is greater than expected by chance, while in other semantic networks, \textit{e.g.}, the English `Has-A' network, the clustering coefficient is smaller than expected by chance.

We hypothesize that these semantic networks are organized according to different principles. It is commonly known that one such principle is similarity: all factors being equal, similar nodes are more likely to be connected. Similarity is believed to play a leading role in the formation of ties in social networks and lies at the heart of many network inference methods. At the same time, recent works indicate that many networks may be organized predominantly according to the complementarity principle, which dictates that interactions are preferentially formed between nodes with complementary properties. Complementarity has been argued to play a significant role in protein-protein interaction networks~\cite{kovacs2019network} and production networks~\cite{mattsson2021functional}. In addition, a geometric complementarity framework for modeling and learning complementarity representations of real networks was recently formulated by Budel~and~Kitsak~\cite{kitsak2020latent}.

This section aims to assess the relative roles of complementarity and similarity mechanisms in different semantic networks. We utilize the method by Talaga~and~Nowak~\cite{talaga2022}. The method assesses the relative roles of the two principles by measuring the relative densities of triangular and quadrangle motifs in the networks. Intuitively, the transitivity of similarity - $A$ similar to $B$ and $B$ similar to $C$ implies $A$ similar to $C$ -- results in a high density of triangles~\cite{kossinets2009origins,rivera2010dynamics,asikainen2020cumulative}, Supplementary Figure~\ref{fig:motifs}a. The non-transitivity of complementarity, on the other hand, suppresses the appearance of triangles but enables the appearance of quadrangles in networks~\cite{kovacs2019network,jia2021measuring}.

We measure and compare the density of triangles and quadrangles with the structural similarity and complementarity coefficients using the framework of Talaga and Nowak~\cite{talaga2022}. After computing the densities of triangles and quadrangles, the framework assesses their significance by comparing the densities to those of the configurational models built with matching degree distributions, see the SI for a summary. As a result of the assessment, the network of interest is quantified by two normalized structural coefficients corresponding to complementarity and similarity. 

\begin{figure}[h]
	\centering
	\includegraphics[width=\textwidth]{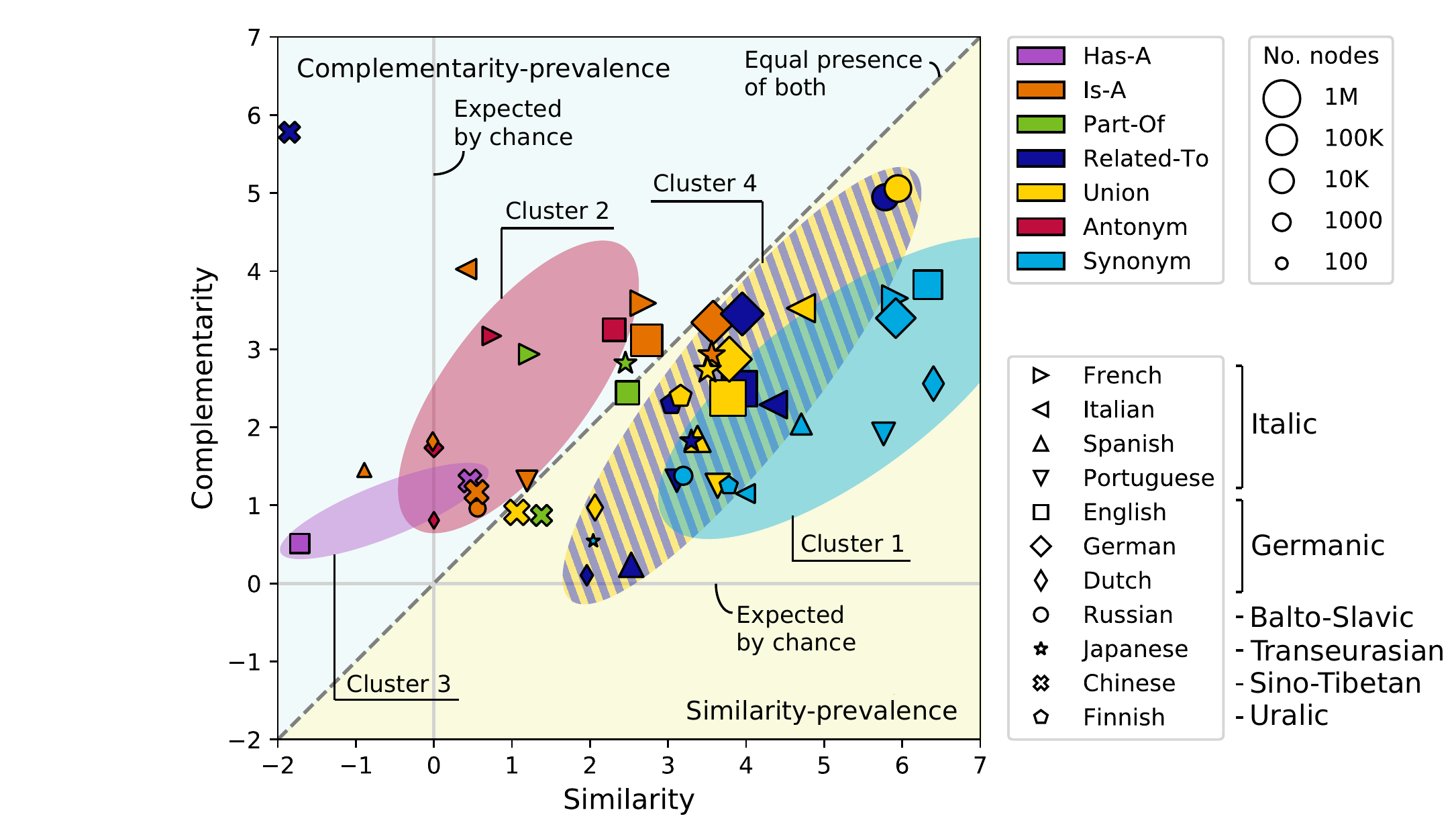}
	\caption{Calibrated average structural coefficients of the LCCs of the 50 semantic networks from 11 languages. Languages that belong to the same family are marked with similar shapes. Triangles represent Italic, quadrilaterals represent Germanic, circles represent Balto-Slavic, a star represents Transeurasian, a cross represents Sino-Tibetan and a pentagon represents Uralic. The marker size scales logarithmically with the number of nodes $N$ in the network and is further adjusted for visibility. The grey lines at $x=0$ and $y=0$ indicate the expected coefficients based on the configuration model (see SI). The dashed line at $y=x$ indicates that the structural similarity and complementarity coefficients are equal. Networks in the upper left area (shaded in blue) are more complementarity-based, while networks in the lower right area (shaded in yellow) are more similarity-based. We highlight four clusters of networks using different colors.}
	\label{fig:similcompl}
	\end{figure}

Figure~\ref{fig:similcompl} depicts the relative roles of complementarity and similarity in 50 semantic networks. We observe that semantic networks are clustered together according to semantic relation types and not their language families, indicating that specific types of semantic relations matter more for the organizing principles of a semantic network rather than its language. 

Based on the calibrated complementarity and similarity values, we can categorize most semantic networks as (i) predominantly complementarity-based, (ii) predominantly similarity-based, and (iii) networks where both complementarity and similarity are substantially present.

We observe four clustering patterns in Figure~\ref{fig:similcompl}.

% Patterns
\begin{itemize}
\item Cluster 1 (light blue): the `Synonym' networks are characterized by stronger similarity than complementarity values. This observation is hardly surprising since `Synonym' networks link words with similar meanings. Since similarity is transitive, the Synonym' networks contain significant numbers of connected node triples, leading to large clustering coefficients.

\item Cluster 2 (red): the `Antonym' networks, as observed in Fig.~\ref{fig:similcompl}, belong to the upper triangle of the scatter plot plane, hence complementarity is more prevalent in these networks than similarity. This observation is as expected, as antonyms are word pairs with opposite meanings that complement each other. In our earlier work~\cite{kitsak2020latent} we learned a geometric representation of the English `Antonym' network demonstrating that antonyms indeed complement each other.

More surprisingly, some antonym networks are characterized by substantial similarity values, implying the presence of triangle motifs. This is the case since there are instances of three or more words that have opposite meanings to all other words in the group. One example is the triple of words (\textit{man}, \textit{woman}, \textit{girl}). Each pair of words in the triple is opposite in meaning along a certain dimension, here either gender or age. 

\item Cluster 3 (purple): the `Has-A' networks show more complementarity than similarity. Intuitively, words in `Has-A' complement one another. For instance, `a \textit{house} has a \textit{roof}' describes a complementary relation and these two objects are not similar to one another.

\item Cluster 4 (dark blue/yellow): Most of the `Related-To' and `Union' networks show more similarity, except for Chinese. As defined in the `Related-To' relation, words are connected if there is any sort of positive relationship between them, therefore triangles are easily formed. One exception to that rule is that the Chinese `Related-To' network (dark blue cross) shows the strongest complementarity of all networks and lower-than-expected similarity. We find that a possible explanation is that the Chinese language has many measure words that are connected to a wide range of nouns. Measure words, also known as numeral classifiers, are used in combination with numerals to describe the quantity of things~\cite{tai1994chinese, cheng1998yi}. For example, the English `one apple' translates to \includegraphics{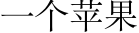} in Chinese, where the measure word \includegraphics{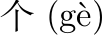} must be added as a unit of measurement between the number `one', \includegraphics{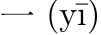}, and the noun `apple', \includegraphics{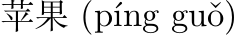}. The Chinese measure word \includegraphics{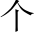} can be loosely translated to English as `unit(s) of', as in `one unit of apple'. This grammatical construct is comparable to the phrase `one box of apples' in English, where `box' serves as a measure word, but, contrary to Chinese, measure words are rare in English. In the Chinese `Related-To' network, there are many measure words that are connected to multiple nouns, and these nouns may have no connection with each other at all. Most of the measure words are not connected with each other either. Hence, the pairings of measure words and nouns lead to quadrangles, a likely explanation of why the Chinese `Related-To' network shows the highest complementarity.

\end{itemize}

%%% DISCUSSION %%%%%%%%%%%%%%%%%%%%%%%%%%%%%%%%%%%%%%%%%%%%%
\section*{Discussion}\setcurrentname{Discussion}\phantomsection\label{sec:discussion}

In summary, we have conducted an exploratory analysis of the topological properties of semantic networks with 7 distinct semantic relations arising from 11 different languages. We identified both universal and unique characteristics of these networks.

We find that semantic networks are sparse and that many are characterized by a power-law degree distribution. We also find that many semantic networks are scale-free. We observe two different patterns of degree-degree mixing in these networks, some networks are assortative, while some are disassortative. In addition, we find that most networks are more clustered than expected. 

On the other hand, some semantic networks -- `Related-To' in French, Spanish, Portuguese, and Finnish -- have unique features that can be explained by rules of grammatical inflection. Because of the grammar in these languages, words have many conjugations or declensions. We have related anomalous peaks in the degree distributions to the language inflections. Notably, we found word inflection not only in inflecting languages but also in Finnish, which is an agglutinating language.  

We have also quantified the relative roles of complementarity and similarity principles in semantic networks. The proportions of similarity and complementarity in networks differ depending on the semantic relation type. For example, the `Synonym' networks exhibit stronger similarity, while the links in the `Antonym' network are primarily driven by complementarity. In addition, the Chinese `Related-To' network has the highest structural complementarity coefficient of all networks, which we attribute to a unique grammatical phenomenon in Chinese: measure words.

\begin{figure}[h]
	\centering
	\includegraphics[width=0.9\textwidth]{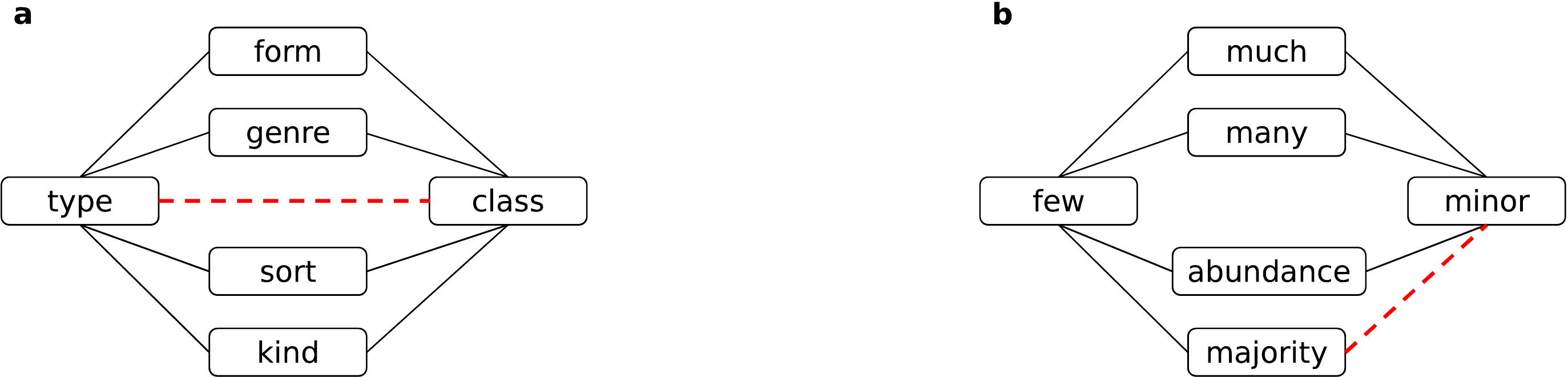}  
	\caption{Examples of similarity and complementarity in semantic networks. (a) Similarity: triangle closure in the `Synonym' network. (b) Complementarity: quadrangle closure in the `Antonym' network.}
	\label{fig:linkpredict}
	\end{figure}

Through the analysis of the topological properties of semantic networks, we found that complementarity may play an important role in their formation. Since most of the state-of-the-art network inference methods are either built on or inspired by the similarity principle, we call for a careful re-evaluation of these methods when it comes to inference tasks on semantic networks. One basic example is the prediction of missing links. In a seminal work, Kov\'{a}cs~\textit{et al.}~\cite{kovacs2019network} demonstrated that protein interactions should be predicted with complementarity-tailored methods. We expect that similar methods might be in place for semantic networks. Instead of using the triangle closure principle, one might benefit from the methods based on quadrangle closure, Figure~\ref{fig:linkpredict}.

It is not as easy to illustrate quadrangle closure in network embedding methods or NLP methods in general. A plethora of methods use multiple modules and parameters in learning tasks and can, in principle, be better optimized for the complementarity structure of semantic networks. In our recent work, we propose a complementarity learning method and apply it to several networks, including the `Antonym' semantic network~\cite{kitsak2020latent}.

Recent groundbreaking advances in large language models are attributed to the multi-head attention mechanism of the Transformer, which uses ideas consistent with the complementarity principle~\cite{vaswani2017attention}. We advocate that a better understanding of statistical mechanisms underlying semantic networks can help us improve NLP methods even further.

\section*{Methods}\setcurrentname{Methods}\phantomsection\label{sec:methods}

\subsection*{Clustering coefficient}
We investigate clustering in semantic networks by measuring the clustering coefficient $c_G(i)$ of a node $i$, which equals the ratio of the number $y$ of connected neighbors to the maximum possible number of connected neighbors,
\begin{equation}
    c_{G}(i) = \dfrac{2y}{d_{i}(d_{i}-1)}, \label{eq:clust_coeff}
\end{equation}
as defined by Watts and Strogatz~\cite{watts1998collective,VanMieghem2014}. The graph clustering coefficient $c_G$ is the average over all node clustering coefficients,
\begin{equation}
    c_{G} = \frac{1}{N}\sum_{i\,=\,1}^{N} c_{G}(i).
\end{equation}
We also calculate the average clustering coefficient $c_G(i)$ of nodes with degree $d_{i}=k$. In addition, we calculate $c_G(i)$ also after random rewiring for comparison.

\section*{Data availability statement}
This study did not generate any new data. Networks used in this study are freely available from \href{https://conceptnet.io}{https://conceptnet.io}.

\bibliography{reference}

\section*{Acknowledgements}

We thank R. Kooij for useful discussions and suggestions. 

This work is part of NExTWORKx, a collaboration between TU Delft and KPN on future telecommunication networks. Parts of this research have been funded by the European Research Council under the European Union’s Horizon 2020 research and innovation program (Grant Agreement 101019718) and the Dutch Research Council (NWO) grant OCENW.M20.244.

\section*{Author contributions statement}

M.K. conceived the experiments, Y.J. conducted the experiments with support from G.B. and M.K, after which G.B., Y.J., P.V. and M.K. analyzed the results. G.B. and Y.J. edited the manuscript and created the visualizations. All authors reviewed the manuscript. 

\section*{Competing interests}
The authors declare no competing interests.

%\newpage

%%% SUPPLEMENTARY INFORMATION %%%%
\supplementarysection

\flushbottom
\maketitle

\subsection*{Data}\setcurrentname{Data}\phantomsection\label{sec:SI_data}
ConceptNet~\cite{ConceptNet} is a multilingual database in the form of a semantic network where nodes are words and phrases from natural language. The links indicate in total 34 different semantic relations. The `knowledge' is collected from a variety of resources, including crowdsourced resources, expert-created resources, and games with a purpose~\cite{ConceptNet}. We study the semantic networks from ConceptNet as it is one of the richest semantic network resources available.

We study the networks belonging to in total 7 link types (relations), 6 of which are directly contained within ConceptNet. They are the `Has-A', `Part-Of', `Is-A', `Related-To', `Antonym' and `Synonym' relations. These are the relation types in ConceptNet that we deem the most meaningful and which also have a sufficient amount of data. In addition, we define an additional link type `Union', which is the set union of the nodes and links of four networks: `Has-A', `Part-Of', `Is-A' and `Related-To'. The purpose of adding this link type is to treat all four relations equally and to evaluate how the structure of the whole network is different from the individual ones. The definitions of the six selected relations from ConceptNet and related examples are outlined in Table~\ref{tab:defrels}. The links of some networks are directed, \textit{i.e.}, of the `Has-A', `Part-Of', and `Is-A' networks, but we treat all networks as undirected for simplicity of the analysis and also comparability. We remove nodes with phrases consisting of more than 5 words, as we deem these to be artifacts of the automated data extraction in ConceptNet.

\subsection*{Semantic relations}
\begin{table}[h]
	\small
	\centering
	\begin{tabular}{C{1.5cm} p{6.5cm} c C{3.0cm} C{2.8cm}}
	\toprule
	 Relation    & \textit{\textbf{Description}}   & \textit{\textbf{Directed}}   & \textit{\textbf{Example}}   & \textit{\textbf{Creation Method}} \\
	\hline
	 \textit{Has-A}       & B belongs to A, either as an inherent part or due to a social construct of possession. Has-A is often the reverse of Part-Of.     & Yes         & bird $\rightarrow$ wing   &   Manual + Automatic    \\ \hline
	 \textit{Part-Of}     & A is a part of B. This is the part meronym relation in WordNet.   & Yes &  gearshift $\rightarrow$ car  & Manual + Automatic     \\ \hline
	 \textit{Is-A}      & A is a subtype or a specific instance of B; every A is a B. This can include specific instances; the distinction between subtypes and instances is often blurry in language. This is the hyponym relation in WordNet.            & Yes          & car $\rightarrow$ vehicle  & Manual + Automatic   \\ \hline
	 \textit{Related-To}     & The most general relation. There is some positive relationship between A and B, but ConceptNet can't determine what that relationship is based on the data.  & No  & learn $\leftrightarrow$ erudition  & Manual + Automatic\\ \hline
	 \textit{Antonym}  & A and B are opposites in some relevant way, such as being opposite ends of a scale, or fundamentally similar things with a key difference between them. Counterintuitively, two concepts must be quite similar before people consider them antonyms. This is the antonym relation in WordNet.  & No & black $\leftrightarrow$ white & Automatic\\ \hline
	 \textit{Synonym} & A and B have very similar meanings. They may be translations of each other in different languages. This is the synonym relation in WordNet. & No & sunlight $\leftrightarrow$ sunshine & Automatic\\
	\bottomrule
	\end{tabular}
	\caption{Definition of the six relations and related information from ConceptNet \cite{Conceptnetrelations}.} \label{tab:defrels}
	\end{table}

\subsection*{Overview of the English semantic networks}
In Table~\ref{tab:fullstats}, we list basic descriptive statistics of the seven semantic networks: the number of nodes $N$, the number of links $L$, the maximum degree $d_{max}$ and the average degree $E[D]$. 

Based on the number of nodes, network `Has-A' is the smallest ($N~=~7,503$) and network `Union' is the largest ($N~=~677,426$). The number of links $L$ ranges from 5,421 to 1,819,646. Relative to the network sizes, all 7 networks have a small average degree, ranging from 1.45 to 5.43. For instance, in network `Part-Of', on average a node only has connections to 2 (0.02\%) of the in total 11,839 nodes. In other words, the number of links is of the same order as the number of nodes, which indicates that semantic networks are sparse. 
	\begin{table}[h]
	\small{\tiny }
	\centering
	\begin{tabular}{lrrrrrrr}
	\toprule
	 Network                                  & \textbf{\textit{Has-A}}   & \textbf{\textit{Is-A}}   & \textbf{\textit{Part-Of}}   & \textbf{\textit{Related-To}}   & \textbf{\textit{Union}}   & \textbf{\textit{Antonym}}   & \textbf{\textit{Synonym}}   \\
	\midrule
	 $N$                                      & 7,503            & 152,538         & 11,839             & 592,816               & 677,426          & 16,867             & 166,922            \\
	 $L$                                      & 5,421            & 220,589         & 12,003             & 1,610,452             & 1,819,646        & 14,371             & 155,048            \\
	 $d_{max}$                                & 372              & 2913            & 116                & 4025                  & 5263             & 38                 & 103                \\
	 $E[D]$                                   & 1.45             & 2.89            & 2.03               & 5.43                  & 5.37             & 1.70               & 1.86               \\
	\bottomrule
	\end{tabular}
	\caption{Basic statistics of the seven English semantic networks extracted from ConceptNet.}
	\label{tab:fullstats} 
	\end{table}

 \subsection*{Descriptive statistics semantic networks}
Table~\ref{tab:statslcc} shows the overall descriptive statistics of the English semantic networks: the number of nodes $N$, the number of links $L$, the maximum degree $d_{max}$, the average degree $E[D]$, the average nearest neighbor degree (ANND), the graph clustering coefficient $c_G$ and the estimated power-law exponents $\hat{\gamma}$. We rewire all semantic networks using the methods described before, after which the same statistics are calculated for the rewired networks.

For networks obtained by degree-preserving rewiring, only the ANND and the graph clustering coefficient $c_G$ change. The average nearest neighbor degree ANND becomes smaller for all randomized semantic networks, except for the `Synonym' network. 

All networks except the `Has-A' network have a remarkably larger graph clustering coefficient $c_G$ (at least by an order of magnitude) than the randomized networks. Because in random networks links are randomly distributed, there are fewer triangles. On the contrary, the randomized networks of `Has-A' exhibit a clustering coefficient more than seven times larger than their original networks.

	\begin{table}[htbp]
	\small
	\centering
	\begin{tabular}{lrrrrrrr}
	\toprule
	Network & \textbf{\textit{Has-A}}   & \textbf{\textit{Is-A}}   & \textbf{\textit{Part-Of}}   & \textbf{\textit{Related-To}}   & \textbf{\textit{Union}}   & \textbf{\textit{Antonym}} & \textbf{\textit{Synonym}} \\
	\midrule
		$N$                                      & 1,664            & 140,024         & 7,562              & 571,079               & 650,079          & 5,912              & 53,279             \\
		$L$                                      & 1,842            & 213,319         & 9,212              & 1,598,548             & 1,804,666        & 7,986              & 80,668             \\
		$d_{max}$                                & 198              & 2913            & 116                & 4025                  & 5263             & 38                 & 103                \\
		$E[D]$                                   & 2.21             & 3.05            & 2.44               & 5.60                   & 5.55            & 2.70                & 3.03               \\
		\textit{ANND}                            & 33.6             & 242             & 14.1               & 170                   & 219              & 6.77               & 7.13               \\
		\textit{ANND rewired}                    & 23.3             & 142             & 10.8               & 145                   & 173              & 6.25               & 7.51               \\
		$c_{G}$                                  & \num{2.17e-03}   & \num{5.66e-02}  & \num{4.61e-02}     & \num{1.02e-01}        & \num{1.04e-01}   & \num{1.50e-02}     & \num{1.13e-01}     \\
		\textit{$c_{G}$ rewired}                 & \num{1.83e-02}   & \num{6.26e-03}  & \num{1.95e-03}     & \num{3.26e-03}        & \num{3.68e-03}   & \num{7.26e-04}     & \num{1.48e-04}     \\
		\textit{$\hat\gamma^{Slope}$} & 2.3  & 2.3  & 2.4  & 2.4  & 2.4  & $\bigtimes$  & $\bigtimes$ \\
        \textit{$\hat\gamma^{Hill}$}  & 2.3  & 2.3  & 2.5  & 2.3  & 2.3  & $\bigtimes$  & $\bigtimes$ \\
        \textit{$\hat\gamma^{Mom}$}   & 2.5  & 2.3  & 2.6  & 2.2  & 2.2  & $\bigtimes$  & $\bigtimes$ \\
        \textit{$\hat\gamma^{Kern}$}  & 2.6  & 2.3  & 2.7  & 2.1  & 2.1  & $\bigtimes$  & $\bigtimes$ \\
	\bottomrule
	\end{tabular}
	\caption{Statistics of the LCCs of seven English semantic networks extracted from ConceptNet. A cross ($\bigtimes$) indicates the degree sequence of the corresponding network is hardly or no power-law.}%%  
	\label{tab:statslcc}
	\end{table}

In summary, we find universalities across semantic networks from different languages in the \textit{degree distribution}, \textit{degree assortativity}, \textit{clustering}, \textit{sparsity} and \textit{connectedness}. Most semantic networks have power-law degree distributions and most of them are scale-free networks. There are two types of degree mixing patterns in semantic networks: assortative and disassortative. Most networks have higher average clustering coefficients than expected by chance, except for one network, the network `Has-A', which shows lower clustering. All semantic networks have high sparsity. Most networks have a single connected component containing the majority of the nodes, except for the network `Has-A', which is more fragmented.
%%%%%%%%%%%%%%%%%%%%%%%%%%%%%%%%%%%%%%%%%%%%%%%%%%%%%%%%%%%%%%%%%%%%%

 \subsection*{Number of nodes in the LCCs of the English networks}
	\begin{table}[H]
	\small \centering
	\begin{tabular}{l|l l l}
	\toprule
	 Network     &Size of full network  & Number of nodes in LCC &  Percentage\\
	\midrule
	 \textit{Has-A}               &\multirow{2}{*}{7,503} &1,664          &  22.18\%   \\
	 \textit{\textcolor{blue}{Has-A (rewired)} }    & &\textcolor{blue}{2,416 $\pm$ 35}     &  \textcolor{blue}{(32.20 $\pm$ 0.47)\%}   \\\hline \rule{0pt}{12pt}
	 \textit{Is-A}                &\multirow{2}{*}{152,538} &140,024   & 91.80\%    \\
	 \textcolor{blue}{\textit{Is-A (rewired)}}      & &\textcolor{blue}{127,258 $\pm$ 73}   & \textcolor{blue}{(83.43 $\pm$ 0.05)\%}    \\\hline \rule{0pt}{12pt}
	 \textit{Part-Of}             &\multirow{2}{*}{11,839} &7,562     & 63.87\%    \\
	 \textcolor{blue}{\textit{Part-Of (rewired)}}   & &\textcolor{blue}{7,993 $\pm$ 53}    & \textcolor{blue}{(67.51 $\pm$ 0.45)\%}    \\\hline \rule{0pt}{12pt}
	 \textit{Related-To}          &\multirow{2}{*}{592,816} &571,079   &  96.33\%   \\
	 \textcolor{blue}{\textit{Related-To (rewired)}}& &\textcolor{blue}{570,012 $\pm$ 116}   &  \textcolor{blue}{(96.15 $\pm$ 0.02)\%}   \\\hline \rule{0pt}{12pt}
	 \textit{Union}               &\multirow{2}{*}{677,426} &650,079   & 95.96\%   \\
	 \textcolor{blue}{\textit{Union (rewired)}}     & &\textcolor{blue}{650,474 $\pm$ 182}   & \textcolor{blue}{(95.77 $\pm$ 0.03)\%}   \\\hline \rule{0pt}{12pt}
	 \textit{Antonym}             &\multirow{2}{*}{16,867} &5,912     &  35.05\%   \\
	 \textcolor{blue}{\textit{Antonym (rewired)} }  & &\textcolor{blue}{8,845 $\pm$ 59}     &  \textcolor{blue}{(52.44 $\pm$ 0.35)\%}   \\ \hline \rule{0pt}{12pt}
	 \textit{Synonym}             &\multirow{2}{*}{166,922} &53,279     &  31.92\%   \\
	 \textcolor{blue}{\textit{Synonym (rewired)}}   & &\textcolor{blue}{103,466 $\pm$ 142}     &  \textcolor{blue}{(61.98 $\pm$ 0.09)\%}   \\
	\bottomrule
	\end{tabular}
	\caption{Number of nodes in the LCCs of the seven English networks in the original and rewired networks. The LCC sizes of the rewired networks are each the average over 10 rewiring realizations with standard deviation shown.} 
	\label{tab:lcc}
	\end{table}

\subsection*{Degree-preserving network rewiring} 
\begin{figure}[h]
	  \centering
	  \includegraphics[width=0.6\textwidth]{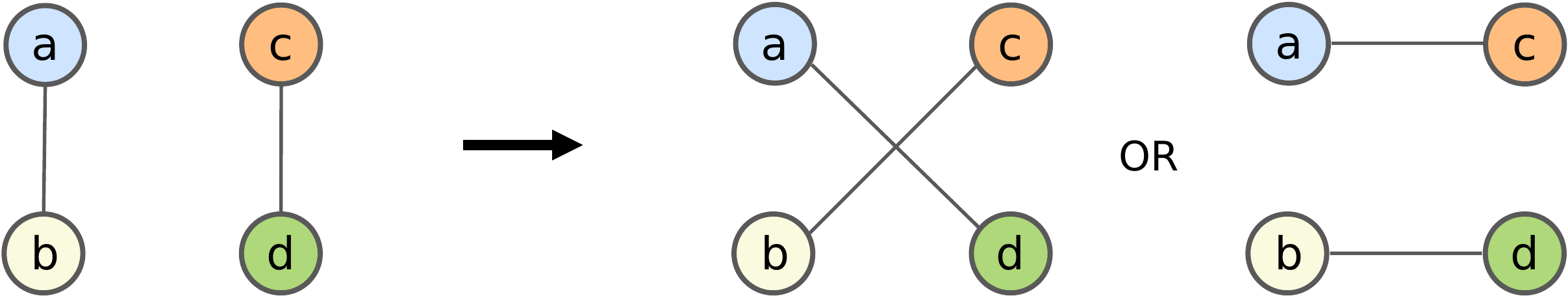}
	\caption{Illustration of degree-preserving rewiring. By randomly swapping the endpoints of two links $(a,b)$ and $(c,d)$, new links can be constructed without changing the node degrees.}
	\label{fig:rewire}
\end{figure}

Degree-preserving network rewiring randomly rewires the links between nodes without changing the node degrees. To preserve the degrees of all nodes, we randomly select 1 link pair (4 nodes) and swap the endpoints of these 2 links. Figure~\ref{fig:rewire} illustrates the rewiring method. To make sure that all links are likely to be rewired at least once, we repeat the random selection of links for $T$ times, where we choose $T=4L$, four times the number of links. The pseudocode is provided in Algorithm~\ref{alg:rewire}.

\subsection*{Distribution of connected components}
\begin{figure}[H]
    \centering
    \includegraphics[width=\textwidth]{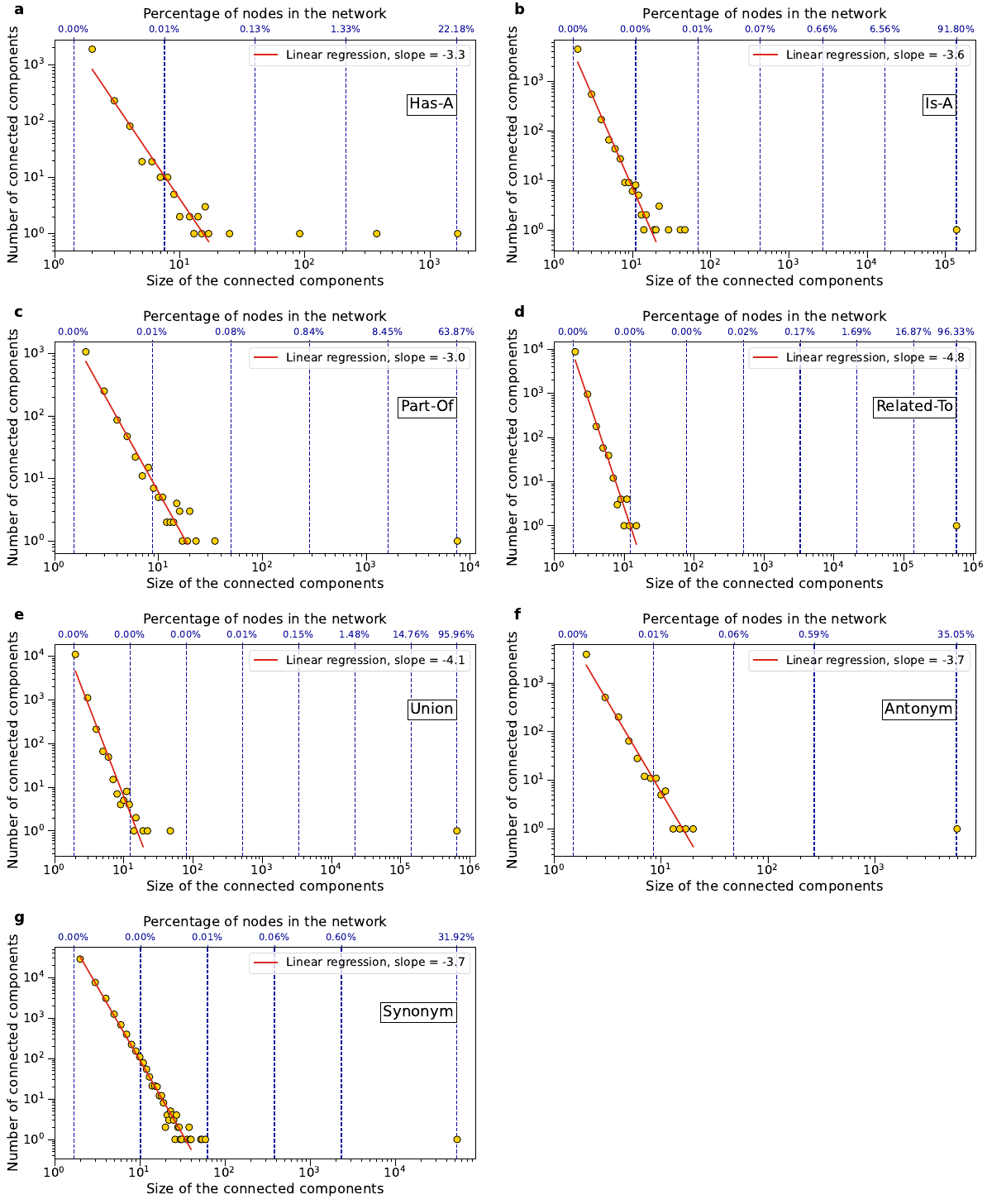}
    \caption{Size distributions of connected components of the seven English semantic networks. The dashed lines indicate the percentage of nodes in the connected components relative to the network total. The regression line is fitted on the bulk of the data points.}
    \label{fig:ccsize}
\end{figure}

We compute all connected components for each network and count the occurrence of the different component sizes. The results are presented in Figure~\ref{fig:ccsize}. Overall, almost every network has a large connected component that is several orders of magnitude larger than the other connected components, except for the network `Has-A', which has multiple larger connected components. Hence, network `Has-A' is more fragmented, having three relatively larger connected components, where the node with the largest degree is not in the LCC but in the second largest one. We inspected each of these three connected components and find that each of the components has a distinct theme. For example, the component with the largest degree node contains all kinds of disease names. We believe that the fragmentation is caused by the partial automatic creation of the dataset.

	\RestyleAlgo{ruled}
	\begin{algorithm}[H]
	\caption{Degree-preserving network rewiring}\label{alg:rewire}
	\KwData{a list of links}
	\KwResult{a rewired network}
    $E \gets$ a list of links\;
	$T \gets 4L$\  \Comment*[r]{all links are rewired at least once}
	\While{$T \neq 0$}{
		$(a, b)$ and $(c, d) \gets$ randomly pick 2 links from $E$\; 
		$n \gets |\text{set}(a,b,c,d)|$  \Comment*[r]{number of unique nodes in 2 links}
		  \eIf{$n < 4$}{
		    \Continue
		  }{$ a$ and $c \gets$ randomly select one node from each link \; 
		    $(c, b) $ and $(a,d) \gets$ swap the two selected nodes\;
		    \eIf{$(c, b) \in E$ or $(a,d) \in E$}
		     {\Continue}
		     {$E \gets$ update the list of links with the 2 rewired links $(c, b) $ and $(a,d)$\;
		     $T \gets T-1$
		     }
		   }
	}
	\end{algorithm}

\newpage

\subsection*{Power-law degree distributions}\setcurrentname{Power-law degree distributions}\phantomsection\label{sec:power_laws}

\subsubsection*{Logarithmic binning}\label{logbinng}
To suppress noise at larger values of the degree $k$ in the density of the degree distribution $\Pr[D=k]$ (the tail), we group the data in bins of equal logarithmic width. In linear binning, every bin has the same linear width $w = k_{i+1}-k_i$, while in logarithmic binning, the bins have constant logarithmic width $b$, where $b=\log(k_{i+1})-\log(k_i)$ \cite{logbinning}. Thus, the linear bin width of a logarithmic bin, $w_i=k_{i+1}-k_i=k_i(e^b-1)$, is proportional to $k_i$. The sizes of the logarithmic bins grow exponentially. Therefore, the number of observations $x$ in a bin is equal to the density of observations $f(k)$ in that bin times the width $w$ of that bin. 

\subsubsection*{Simple power-law exponent estimation}
A common method for estimating the power-law exponent $\gamma$ in $\Pr[D=k] \approx c k^{-\gamma}$ is to measure the slope of $\log(\Pr[D=k])$ versus $\log(k)$. Since the probability density function $f(k)$ of the degree is proportional to $k^{-\gamma}$, the number of observations $x \propto f(k)\times w \propto k^{1-\gamma}$. Regressing $\log(x)$ against $\log(k)$ yields a slope equal to $1-\gamma$. To estimate $\gamma$, normalization of the number of observations $x$ is required. Due to the increasing bin width, a bin can contain more than one value of $k$. The sum of all observations within a bin is $x$. To preserve the probability of a node with degree $k$ such that the total probability of degree distribution is equal to 1, the number of observations $x$ is normalized by the linear width of the bin. This converts $x$ to the number of observations per unit of the bin width, $(x/w)\propto k^{-\gamma}$. As a result, regressing the normalized logarithmic bin counts $\log{\left(x/w\right)}$ against the logarithmic degree $\log{(k)}$ yields a slope of $-\gamma$ \cite{logbinning}. We base our slope estimate of $\gamma$ on the linear part of the tail in the density $\Pr[D=k]$, which we determine by inspection for each network.

\subsubsection*{Consistent power-law exponent estimators}\setcurrentname{Consistent power-law exponent estimators}\phantomsection\label{sec:consistent_powerlaw}
A more rigorous approach to power-law degree distributions is provided by Voitalov~\textit{et~al.}~\cite{voitalov2019scale}, here we summarize the main conclusions for the convenience of the reader. They consider a degree distribution to be a power law if the probability density function is a member of the class of regularly varying functions: $\Pr[D=k] = \ell(k) k^{-\gamma}$, where $\ell(k)$ is a slowly varying function. The function $\ell(k)$ is called slowly varying if 
\begin{equation}
    \lim_{k\,\rightarrow\,\infty} \dfrac{\ell(ak)}{\ell(k)} = 1,
\end{equation}
for any $a > 0$. This definition corresponds with a perfect power law in the tail of the distribution. Voitalov~\textit{et~al.} propose to use three different consistent estimators of the power-law exponent: the Hill, moments and kernel estimators. We use their software package to obtain these estimates for the degree sequences of our networks~\cite{voitalov2019code}. These estimators do not estimate $\gamma$ directly, but rather the extreme value index 
\begin{equation}
    \xi = \dfrac{1}{\gamma - 1}.
\end{equation}
As a rule of thumb, Voitalov~\textit{et~al.} consider a distribution to be a power law if $\hat{\xi} > 1/4$ for all three estimators~\cite{voitalov2019scale}, corresponding with $\hat{\gamma} < 5$. Here, we adopt this rule for distinguishing power-law distributions. In addition, they call a distribution \textit{hardly power-law} if all $\hat{\xi} > 0$, but at least one $\hat{\xi} \leq 1/4$ ($\hat{\gamma} > 5$). If any $\hat{\xi} < 0$, the distribution is not a power law.

\subsection*{Average nearest neighbor degree as a function of the degree}

\subsection*{Clustering coefficient as a function of degree}

%%%%%%%%%%%%%%%%%%%%%%%%%%%%%%%%%
\subsection*{Descriptive statistics of the semantic networks from different languages}
This section shows the descriptive statistics of semantic networks from the eleven languages. Each property is compared among the seven networks for the eleven languages.

\subsubsection*{Language classifications}
Table~\ref{tab:lang_class} shows the typological and genetic classifications of the eleven considered languages.

\begin{table}[H]
	\centering
	\footnotesize
	\begin{tabular}{lp{2.5cm}<{\centering}cc}
	\toprule
	\diagbox{Genetic}{Typological} & \textit{\textbf{Inflecting}} & \textit{\textbf{Isolating}} & \textit{\textbf{Agglutinating}}\\ 
        \midrule
        \textit{Italic} & Spanish, French, Italian, Portuguese & &	\\ \midrule
        \textit{Germanic} & English, Dutch, German & &	\\\midrule
        \textit{Balto-Slavic} & Russian & &	\\\midrule
        \textit{Transeurasian} & & & Japanese \\\midrule
        \textit{Sino-Tibetan} & & Chinese &	\\\midrule
        \textit{Uralic} & & & Finnish	\\      
	\bottomrule
	\end{tabular}
	\caption{Genetic and typological language classifications of the eleven languages.}
	\label{tab:lang_class}
	\end{table}

\subsubsection*{Overview statistics of semantic networks from different languages}
Table~\ref{tab:langs_nodes} shows the number of nodes of each semantic network in the eleven different languages. A blank element in the table indicates that the network does not exist, \textit{i.e.}, a relation is not available in that language. 

	\begin{table}[H]
	\small
	\centering
	\begin{tabular}{lrrrrrrr}
	\toprule
	Network & \textbf{\textit{Has-A}}   & \textbf{\textit{Is-A}}   & \textbf{\textit{Part-Of}}   & \textbf{\textit{Related-To}}   & \textbf{\textit{Union}}   & \textbf{\textit{Antonym}} & \textbf{\textit{Synonym}} \\
	\midrule
	 \textit{English}    & 1,664            & 140,024         & 7,562              & 571,079               & 650,079          & 5,912              & 53,279             \\
	 \textit{French}     &                  & 17,519          & 2,832              & 1,289,083             & 1,296,622        & 1,361              & 20,144             \\
	 \textit{Italian}    &                  & 2,663           & 9                  & 36,295                & 46,468           & 13                 & 1,580              \\
	 \textit{German}     &                  & 113,301         & 5                  & 100,737               & 172,147          & 187                & 43,072             \\
	 \textit{Spanish}    &                  & 255             & 11                 & 12,094                & 22,861           & 15                 & 3,491              \\
	 \textit{Russian}    &                  & 557             & 3                  & 20,268                & 25,887           & 12                 & 1,148              \\
	 \textit{Portuguese} &                  & 3,341           & 15                 & 5,929                 & 11,426           & 17                 & 6,421              \\
	 \textit{Dutch}      &                  & 191             & 53                 & 303                   & 1,418            & 111                & 11,964             \\
	 \textit{Japanese}   & 38               & 40,256          & 7,230              & 7,200                 & 43,286           & 20                 & 230                \\
	 \textit{Finnish}    &                  & 76              & 12                 & 4,483                 & 6,958            & 24                 & 1,569              \\
	 \textit{Chinese}    & 6,355            & 10,073          & 3,417              & 3,163                 & 17,128           & 4                  & 17                 \\
	\bottomrule
	\end{tabular}
	\caption{Number of nodes $N$ in the LCCs of the semantic networks from the eleven different languages extracted from ConceptNet. A blank element indicates the corresponding network is not available. The `Union' network is the union of four networks (`Has-A', `Is-A', `Part-Of' and `Related-To'). Because we display the LCC sizes, for some `Union' networks, the number of nodes exceeds the sum of the sizes of its four constituent networks.}
	\label{tab:langs_nodes}
	\end{table}

\subsubsection*{Average degree in the LCCs of semantic networks from different languages}
\begin{table}[H]
	\small
	\centering
	\begin{tabular}{lrrrrrrr}
	\toprule
	Network & \textbf{\textit{Has-A}}   & \textbf{\textit{Is-A}}   & \textbf{\textit{Part-Of}}   & \textbf{\textit{Related-To}}   & \textbf{\textit{Union}}   & \textbf{\textit{Antonym}} & \textbf{\textit{Synonym}} \\
	\midrule
	 \textit{English}    &             2.21 &            3.05 &               2.44 &                   5.60 &             5.55 &                2.70 &               3.03 \\
	 \textit{French}     &                  &            2.64 &               2.51 &                  3.44 &             3.46 &               2.45 &               2.81 \\
	 \textit{Italian}    &                  &            2.86 &               2.89 &                   2.20 &             2.27 &               1.85 &               2.54 \\
	 \textit{German}     &                  &            2.75 &                1.60 &                  4.77 &             4.53 &               2.16 &               3.57 \\
	 \textit{Spanish}    &                  &            2.45 &               2.73 &                  2.13 &             2.13 &               1.87 &               2.57 \\
	 \textit{Russian}    &                  &            2.23 &               1.33 &                  4.14 &             3.88 &               1.83 &               2.26 \\
	 \textit{Portuguese} &                  &            2.24 &               2.67 &                  2.49 &             2.65 &                  2.00 &               2.84 \\
	 \textit{Dutch}      &                  &            4.68 &               4.98 &                   2.30 &             2.69 &               2.11 &               3.53 \\
	 \textit{Japanese}   &             2.89 &            4.42 &               4.11 &                  4.34 &             4.79 &                  2.00 &               2.73 \\
	 \textit{Finnish}    &                  &            1.97 &               1.83 &                   2.30 &             2.26 &               1.92 &               2.24 \\
	 \textit{Chinese}    &             3.58 &            3.02 &               3.36 &                  4.06 &             3.78 &                1.50 &               2.24 \\
	\bottomrule
	\end{tabular}
	\caption{Average degree $E[D]$ in the LCCs of the semantic networks from the eleven different languages extracted from ConceptNet. A blank element indicates the corresponding network is unavailable.}
	\label{tab:langs_ED}
	\end{table}

\subsubsection*{Estimated power-law exponents for semantic networks from different languages}
Table~\ref{tab:langs_gamma} lists the estimated power-law exponents $\hat{\gamma}$ for each semantic network in the eleven languages. We consider a network to not have a power-law degree distribution if it is not or hardly power-law according to the method of Voitalov~\textit{et~al.}\cite{voitalov2019scale}.
\begin{table}[H]
	\small
	\centering
	\begin{tabular}{lllllllll}
	\toprule
	Network & $\gamma$ &\textbf{\textit{Has-A}} &\textbf{\textit{Is-A}} &\textbf{\textit{Part-Of}} &\textbf{\textit{Related-To}} &\textbf{\textit{Union}} &\textbf{\textit{Antonym}} &\textbf{\textit{Synonym}} \\
	\midrule
	 \multirow{4}{*}{\textit{English}} 
        &$\hat\gamma^{Slope}$ & 2.3  & 2.3  & 2.4  & 2.4  & 2.4  & $\bigtimes$  & $\bigtimes$ \\
        &$\hat\gamma^{Hill}$  & 2.3  & 2.3  & 2.5  & 2.3  & 2.3  & $\bigtimes$  & $\bigtimes$ \\
        &$\hat\gamma^{Mom}$   & 2.5  & 2.3  & 2.6  & 2.2  & 2.2  & $\bigtimes$  & $\bigtimes$ \\
        &$\hat\gamma^{Kern}$  & 2.6  & 2.3  & 2.7  & 2.1  & 2.1  & $\bigtimes$  & $\bigtimes$ \\\hline
	 \multirow{4}{*}{\textit{French}} 
        &$\hat\gamma^{Slope}$ &    &2.4  &2.3  &$\bigtimes$  &$\bigtimes$  &2.7  &3.1 \\
        &$\hat\gamma^{Hill}$  &    &2.5  &2.5  &$\bigtimes$  &$\bigtimes$  &3.3  &3.6\\
        &$\hat\gamma^{Mom}$   &    &2.5  &2.6  &$\bigtimes$  &$\bigtimes$  &4.5  &3.9\\
        &$\hat\gamma^{Kern}$  &    &2.6  &2.6  &$\bigtimes$  &$\bigtimes$  &4.2  &4.8\\\hline
      \multirow{4}{*}{\textit{Italian}} 
        &$\hat\gamma^{Slope}$ &    &2.3  &   &2.6  &2.6  &   &$\bigtimes$ \\
        &$\hat\gamma^{Hill}$  &    &2.8  &   &2.4  &2.4  &   &$\bigtimes$\\
        &$\hat\gamma^{Mom}$   &    &2.2  &   &2.4  &2.5  &   &$\bigtimes$\\
        &$\hat\gamma^{Kern}$  &    &2.3  &   &2.5  &2.6  &   &$\bigtimes$\\\hline
      \multirow{4}{*}{\textit{German}} 
        &$\hat\gamma^{Slope}$ &    &2.5  &   &2.6  &2.5  &   &3.1 \\
        &$\hat\gamma^{Hill}$  &    &2.2  &   &2.7  &2.9  &   &3.6\\
        &$\hat\gamma^{Mom}$   &    &3.3  &   &2.7  &2.9  &   &3.7\\
        &$\hat\gamma^{Kern}$  &    &2.3  &   &2.9  &2.6  &   &3.9\\\hline
      \multirow{4}{*}{\textit{Spanish}} 
        &$\hat\gamma^{Slope}$ &    &   &   &$\bigtimes$  &$\bigtimes$  &   &$\bigtimes$ \\
        &$\hat\gamma^{Hill}$  &    &   &   &$\bigtimes$  &$\bigtimes$  &   &$\bigtimes$\\
        &$\hat\gamma^{Mom}$   &    &   &   &$\bigtimes$  &$\bigtimes$  &   &$\bigtimes$\\
        &$\hat\gamma^{Kern}$  &    &   &   &$\bigtimes$  &$\bigtimes$  &   &$\bigtimes$\\\hline        
      \multirow{4}{*}{\textit{Russian}} 
        &$\hat\gamma^{Slope}$ &    &   &   &$\bigtimes$  &$\bigtimes$  &   &$\bigtimes$ \\
        &$\hat\gamma^{Hill}$  &    &   &   &$\bigtimes$  &$\bigtimes$  &   &$\bigtimes$\\
        &$\hat\gamma^{Mom}$   &    &   &   &$\bigtimes$  &$\bigtimes$  &   &$\bigtimes$\\
        &$\hat\gamma^{Kern}$  &    &   &   &$\bigtimes$  &$\bigtimes$  &   &$\bigtimes$\\\hline       
      \multirow{4}{*}{\textit{Portuguese}} 
        &$\hat\gamma^{Slope}$ &    &2.6  &   &2.4  &2.5  &   &$\bigtimes$ \\
        &$\hat\gamma^{Hill}$  &    &2.8  &   &2.6  &2.8  &   &$\bigtimes$\\
        &$\hat\gamma^{Mom}$   &    &2.6  &   &2.1  &2.4  &   &$\bigtimes$\\
        &$\hat\gamma^{Kern}$  &    &2.6  &   &2.9  &2.7  &   &$\bigtimes$\\\hline      
      \multirow{4}{*}{\textit{Dutch}} 
        &$\hat\gamma^{Slope}$ &    &   &   &   &2.2  &   &$\bigtimes$ \\
        &$\hat\gamma^{Hill}$  &    &   &   &   &2.8  &   &$\bigtimes$\\
        &$\hat\gamma^{Mom}$   &    &   &   &   &3.1  &   &$\bigtimes$\\
        &$\hat\gamma^{Kern}$  &    &   &   &   &3.5  &   &$\bigtimes$\\\hline        
      \multirow{4}{*}{\textit{Japanese}} 
        &$\hat\gamma^{Slope}$ &    &2.4  &2.3  &2.2  &2.3  &   &  \\
        &$\hat\gamma^{Hill}$  &    &2.6  &2.9  &4.9  &2.6  &   & \\
        &$\hat\gamma^{Mom}$   &    &2.6  &2.9  &2.4  &2.7  &   & \\
        &$\hat\gamma^{Kern}$  &    &2.6  &2.6  &2.6  &2.6  &   & \\\hline       
      \multirow{4}{*}{\textit{Finnish}} 
        &$\hat\gamma^{Slope}$ &    &   &   &$\bigtimes$  &$\bigtimes$  &   &$\bigtimes$ \\
        &$\hat\gamma^{Hill}$  &    &   &   &$\bigtimes$  &$\bigtimes$  &   &$\bigtimes$\\
        &$\hat\gamma^{Mom}$   &    &   &   &$\bigtimes$  &$\bigtimes$  &   &$\bigtimes$\\
        &$\hat\gamma^{Kern}$  &    &   &   &$\bigtimes$  &$\bigtimes$  &   &$\bigtimes$\\\hline          
      \multirow{4}{*}{\textit{Chinese}} 
        &$\hat\gamma^{Slope}$ &2.5   &2.3  &2.7  &1.9  &2.3  &   &  \\
        &$\hat\gamma^{Hill}$  &3.4   &2.4  &2.3  &2.7  &4.3  &   & \\
        &$\hat\gamma^{Mom}$   &3.8   &2.4  &2.4  &1.9  &2.4  &   & \\
        &$\hat\gamma^{Kern}$  &2.7   &2.3  &2.3  &2.3  &2.5  &   & \\\hline 
	\bottomrule
	\end{tabular}
	\caption{Estimated power-law exponents $\hat{\gamma}$ for the LCCs of the semantic networks in different languages. A blank element indicates the corresponding network is either unavailable or the number of nodes $N<1000$. A cross ($\bigtimes$) indicates that the degree sequence of that network is not or hardly power-law.}
	\label{tab:langs_gamma}
	\end{table}

\subsection*{Examples of words in the peak and their neighoring words in the Spanish `Related-To' network}
\begin{table}[H]
	\small
	\centering
	\begin{tabular}{lll}
	\toprule
	Peak word & \textit{\textbf{Translation}} & \textit{\textbf{Neighbors}} \\
	\midrule %\rule{0pt}{12pt}
	\textit{cenar} & to dine & cená, cenábamos, cenáculo, cenáis, cenáramos, cenáremos, ...\\ \midrule %\hline \rule{0pt}{12pt}
	\textit{viajar}& to travel& viaja, viajaba, viajabais, viajaban, viajabas, viajad, viajado, ...\\ \midrule %\hline\rule{0pt}{12pt}
	\textit{pasear}& to walk& pasea, paseaba, paseabais, paseaban, paseabas, pasead, ...\\ \midrule %\hline\rule{0pt}{12pt}
	\textit{reparar}& to repair& repararais, repararan, repararas, reparareis, repararemos, ...\\ \midrule %\hline \rule{0pt}{12pt}
	\textit{comparar}& to compare& comprar, comparaba, comparabais, comparaban, comparabas, ...\\
	\bottomrule
	\end{tabular}
	\caption{Examples of words in the peak and their neighoring words in the Spanish `Related-To' network.}
	\label{tab:es_peak}
	\end{table}

\subsection*{Percentages of POS tags among peak words and in the LCCs of `Related-To' networks in four inflecting languages}
	\begin{table}[H]
	\centering
	\footnotesize
	\begin{tabular}{lrrrrrrrr}
	\toprule
	\multirow{2}{*}{Percentage (\%)}  &  \multicolumn{2}{c}{\textbf{French}} &  \multicolumn{2}{c}{\textbf{Spanish}}   & \multicolumn{2}{c}{\textbf{Portuguese}} & \multicolumn{2}{c}{\textbf{Finnish}}\\ 
					\cmidrule(lr){2-3} \cmidrule(lr){4-5} \cmidrule(lr){6-7} \cmidrule(lr){8-9}
	                    & LCC & Peak & LCC & Peak & LCC & Peak & LCC & Peak\\
	\midrule
	POS tagged & 98.71 & 98.66 &92.72 & 77.84 & 67.60 & 60.00 & 81.37 & 64.13\\ 
	\midrule
	 \textit{Verb}     &  68.90 & 89.97 &87.62 & 98.44 & 32.56 & 100.00 & 11.40 & 11.36 \\
	 \textit{Noun}     &  19.21 & 7.14 & 9.20 & 1.56 & 51.96 & 0 & 77.96 & 84.09 \\
	 \textit{Adjective}&  11.53 & 2.75 & 2.89 & 0 & 14.60 & 0 & 7.17 & 4.55 \\
	 \textit{Adverb}   &  0.36 & 0.15 & 0.29 & 0 & 0.88 & 0 &  3.47 & 0 \\
	\bottomrule
	\end{tabular}
	\caption{Percentages of POS tags among peak words and in the LCCs of the `Related-To' networks of four inflecting languages.}
	\label{tab:wordtypes}
	\end{table}

\subsection*{Percentage of verbs and nouns among the neighbors of peak words of the LCC of `Related-To' networks of four inflecting languages}
	\begin{table}[H]
	\centering
	\footnotesize
	\begin{tabular}{lrrrrrrrr}
	\toprule
	\multirow{2}{*}{Percentage (\%)}  &  \multicolumn{2}{c}{\textbf{French}} &  \multicolumn{2}{c}{\textbf{Spanish}}   & \multicolumn{2}{c}{\textbf{Portuguese}} & \multicolumn{2}{c}{\textbf{Finnish}}\\ 
					\cmidrule(lr){2-3} \cmidrule(lr){4-5} \cmidrule(lr){6-7} \cmidrule(lr){8-9}
	                    & Mean & SD & Mean & SD & Mean & SD & Mean & SD\\
	\midrule
	POS tagged & 97.39 & 0.88 &96.96 & 1.73 & 97.74 & 0.75 & 93.72 & 4.45\\ 
	\midrule
	 \textit{Verb}     &87.26 & 25.85 &97.24 & 2.59 &99.23 & 0.94 & 3.86 & 14.64 \\
	 \textit{Noun}     & 9.34 & 20.08 &2.07 & 2.15 &0.77 & 0.94 & 89.67 & 26.50 \\
	\bottomrule
	\end{tabular}
	\caption{The mean and Standard Deviation (SD) percentage of verbs and nouns in the neighbors of peak words of the LCC of network `Related-To' in four inflecting languages.}
	\label{tab:nb_wordtypes}
	\end{table}

\subsection*{Node merging procedure}
First, we extract the network `Form-Of' in the same way as for all other networks. Then we treat the merged group of words as a single word in the `Related-To' network in the same language. Next, we calculate the number of nodes with degree $k$ in the new `Related-To' network. Finally, we plot the densities of the degree distributions of French, Spanish, Portuguese and Finnish networks. 
\graphicspath{{figures/2results/DiffLangs/}}
	\begin{figure}[H]
	\centering
	\includegraphics[width=0.5\textwidth]{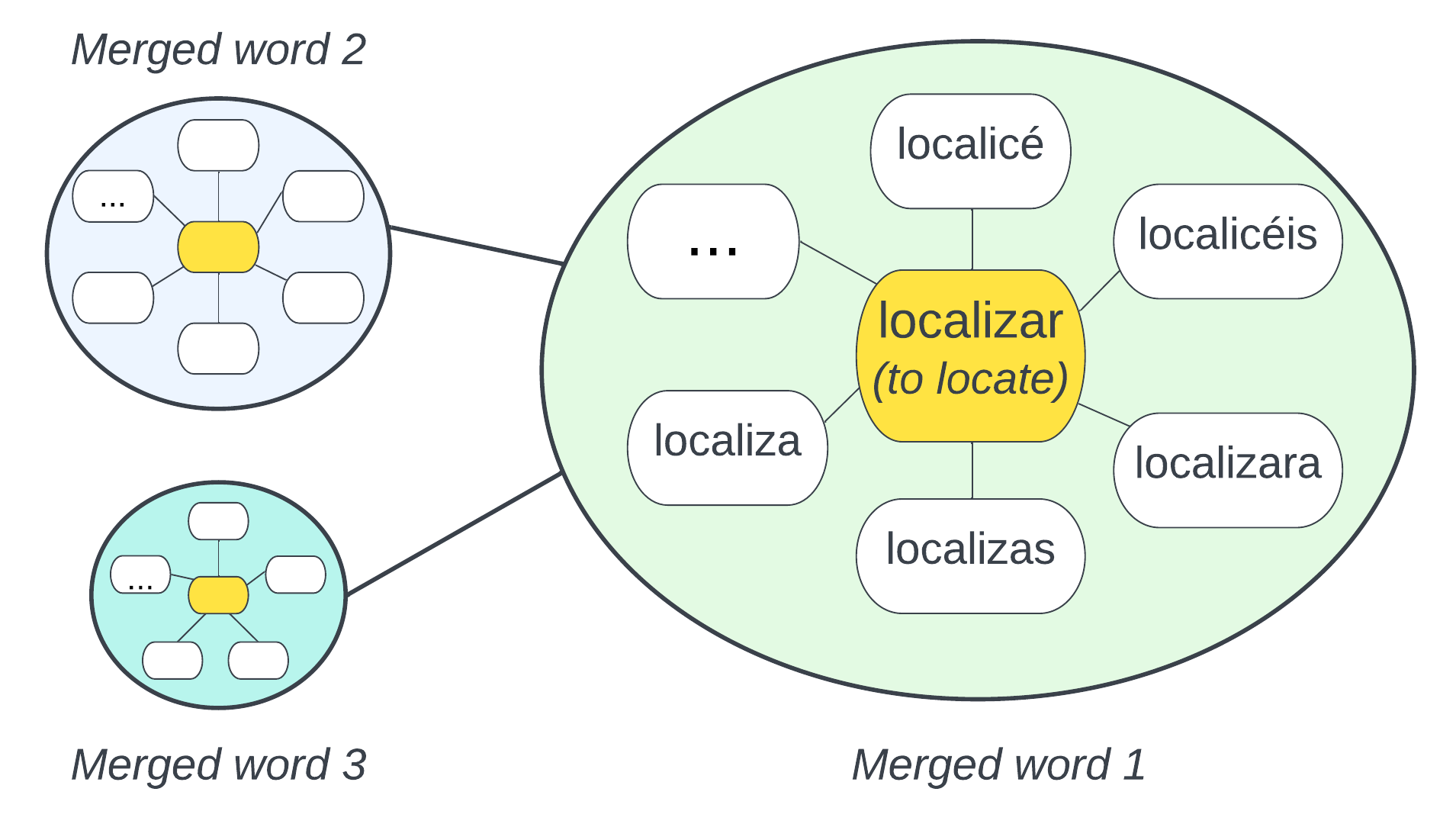}  
	\caption{Illustration of the merging of words in the `Related-To' network. After merging a root word and its neighbors, all words in a circle are seen as a single word.}
	\label{fig:merge}
	\end{figure}

\subsection*{The percentages of matched words among the peak words for the LCCs of the `Related-To' networks in four languages}
	\begin{table}[H]
	\centering
	\small
	\begin{tabular}{p{5cm}rrrr}
	\toprule
	Percentage & \textbf{French}& \textbf{Spanish}& \textbf{Portuguese}& \textbf{Finnish}\\ 
	\midrule
	\textit{Percentage of peak words covered by `Form-Of'} & 33.72\% & 100\% & 60.00\% & 91.30\%  \\ \midrule
    \textit{Percentage of neighbors of peak words covered by `Form-Of'} & 17.38\% & 97.76\% & 55.47\% & 45.08\%\\ 
	\bottomrule
	\end{tabular}
	\caption{The percentages of matched words among the peak words of the LCCs of the `Related-To' networks in four languages.}
	\label{tab:FORT2 commonN}
	\end{table}

\subsection*{The number of grammatical variations in French, Spanish, Portuguese and Finnish}
	\begin{table}[H]
	\centering
	\begin{tabular}{lccc}
	\toprule
	 Language & \textbf{\textit{Grammatical variations m}} & $k_{min}$ & $k_{max}$  \\
	\midrule
	  \textit{French}     & 42 &   36 &     51  \\
	  \textit{Spanish}    & 54 &   45 &     61\\
	  \textit{Portuguese} & 54 &   53 &     53  \\
	  \textit{Finnish}    & 30 &   25 &     35  \\
	\bottomrule
	\end{tabular}
	\caption{The maximum number of grammatical variations $m$ for the grammatical rule of interest in French, Spanish, Portuguese and Finnish. The minimum and maximum degree $k_{min}$ and $k_{max}$ where the peak starts and ends in the densities of the degree distributions of the `Related-To' networks are included for comparison.}
	\label{tab:kmatch}
	\end{table}

%%%%%%%%%%%%%%%%%%%%%%%%%%%%%%%%%%%%%%%%%%%%%%%%%%%%%%%%%
\subsection*{Structural similarity and complementarity coefficients}
\subsubsection*{Structural similarity coefficient}
For the convenience of the reader, here we summarize the main components of the framework for computing structural similarity and complementarity coefficients by Talaga and Nowak~\cite{talaga2022}.

The structural similarity coefficient $s_i$ generalizes the local clustering and closure coefficients. The local clustering coefficient $s_i^W$ of a node $i$ is the classic clustering coefficient. It is defined as the fraction of triples centered at $i$ which can be closed to form a triangle,
	\begin{equation} \label{eq:sw}
	s_i^W=\frac{2 T_i}{t_i^W}=\frac{\sum_{j, k} a_{i j} a_{i k} a_{j k}}{d_i\left(d_i-1\right)},
	\end{equation}
where $T_i$ is the number of triangles including $i$ and $t_i^W$ is the number of wedge triples (Fig.~\ref{fig:motifs}b), or 2-paths with node $i$ in the middle, \textit{e.g.}, $(j,i,k)$. 
The definition of the local closure coefficient \cite{yin2019local} is given as follows
	\begin{equation}\label{eq:sh}
	s_i^H=\frac{2 T_i}{t_i^H}=\frac{\sum_{j, k} a_{i j} a_{i k} a_{j k}}{\sum_j a_{i j}\left(d_j-1\right)},
	\end{equation}
where $t_i^H$ is the number of head triples (Fig.~\ref{fig:motifs}c), \textit{i.e.}, 2-paths starting from node $i$, such as $(i,j,k)$. Both $s_i^W$ and $s_i^H$ are bounded in the range $[0,1]$, but they capture different parts of the spectrum of similarity-driven structures~\cite{talaga2022}.

Combining the weighted average of these two coefficients results in a more comprehensive measure of local structure, the \textit{structural similarity coefficient}~\cite{talaga2022}, which captures the full spectrum of structural similarity. It is defined as 
	\begin{equation}\label{eq:si}
	s_i=\frac{4 T_i}{t_i^W+t_i^H}=\frac{t_i^W s_i^W+t_i^H s_i^H}{t_i^W+t_i^H}.
	\end{equation}
The coefficient $s_i=1$ only if node $i$ is in a fully connected network.

The structural similarity coefficient of a whole network $G$ is then the average over all nodes
	\begin{equation} \label{eq:s}
	s(G)=\frac{1}{N} \sum_{i=1}^{N} s_i.
	\end{equation}

\subsubsection*{Structural complementarity coefficient}
Analogously, the local quadruples clustering coefficient at node $i$ is defined as the fraction of closed quadruples with $i$ at the second position~\cite{talaga2022}
	\begin{equation}\label{eq:cw}
	c_i^W=\frac{2 Q_i}{q_i^W}=\frac{\sum_{j \neq i} a_{i j} \sum_{k \neq i, j} a_{i k}\left(1-a_{j k}\right) \sum_{l \neq i, j, k} a_{k l} a_{j l}\left(1-a_{i l}\right)}{\sum_j a_{i j}\left[\left(d_i-1\right)\left(d_j-1\right)-n_{i j}\right]} ,
	\end{equation}
where $Q_i$ represents the number of quadrangles contain that node $i$ and $q_i^W$ is the number of wedge quadruples (Fig.~\ref{fig:motifs}e), or 3-paths with $i$ at the second node, \textit{e.g.}, $(l,i,j,k)$. 
Similarly, the local quadruples closure coefficient of a node $i$ calculates the percentage of closed quadruples beginning at $i$
	\begin{equation}\label{eq:ch}
	c_i^H=\frac{2 Q_i}{q_i^H}=\frac{\sum_{j \neq i} a_{i j} \sum_{k \neq i, j} a_{i k}\left(1-a_{j k}\right) \sum_{l \neq i, j, k} a_{k l} a_{j l}\left(1-a_{i l}\right)}{\sum_{j \neq i} a_{i j} \sum_{k \neq i, j} a_{j k}\left(d_k-1-a_{i k}\right)} ,
	\end{equation}
where $q_i^H$ is the number of head quadruples originating from node $i$ (Fig.~\ref{fig:motifs}f).

Finally, the \textit{structural complementarity coefficient} is constructed as the weighted average of the local quadruples clustering and closure coefficients \cite{talaga2022}
	\begin{equation}\label{eq:ci}
	c_i=\frac{4 Q_i}{q_i^W+q_i^H}=\frac{q_i^W c_i^W+q_i^H c_i^H}{q_i^W+q_i^H}.
	\end{equation}
The structural complementarity coefficient $c_i \in [0,1]$, which is proven to be a more general measure than using only $c_i^W$ or $c_i^H$ \cite{talaga2022}. The maximum $c_i = 1$ happens only if node $i$ belongs to a fully connected bipartite graph. In a bipartite graph, nodes are divided into two groups, and connections are only formed between groups but not within the same group.

The structural complementarity coefficient of a whole network $G$ is then the average of all nodes:
	\begin{equation}\label{eq:c}
	c(G)=\frac{1}{N} \sum_{i=1}^{N} c_i.
	\end{equation}

Table~\ref{tab:process} lists the procedures of how we compute the structural similarity and complementarity coefficients to quantify the density of triangles
and quadrangles in a network $G$, respectively.
	\begin{table}[H]
	\small
	\centering  
	\begin{tabular}{cl|cc}
	\toprule
	 \multirow{2}{*}{\textit{Procedure}} & \multirow{2}{*}{Structural coefficients} & \multicolumn{2}{c}{Network $G$} \\ \cmidrule(lr){3-4}
	 && \textbf{Similarity} ($\triangle$) & \textbf{Complementarity} ($\square$)\\ \midrule \rule{0pt}{12pt}
	 \multirow{2}{*}{\textit{Step 1}}&Wedge triple/quadruple & $s_i^W$, Eq.~\ref{eq:sw} & $c_i^W$, Eq.~\ref{eq:cw}\\ \rule{0pt}{12pt}
	 & Head triple/quadruple	& $s_i^H$, Eq.~\ref{eq:sh} & $c_i^H$, Eq.~\ref{eq:ch}\\ \hline \rule{0pt}{16pt}
	 \textit{Step 2} & Node-wise 				& $s_i$, Eq.~\ref{eq:si} &  $c_i$, Eq.~\ref{eq:ci} \\ \hline \rule{0pt}{16pt}
	 \textit{Step 3}& Network-wise 			& $s(G)=\frac{1}{N} \sum_{i=1}^{N} s_i$, Eq.~\ref{eq:s} &  $c(G)=\frac{1}{N} \sum_{i=1}^{N} c_i$, Eq.~\ref{eq:c} \\ \hline \rule{0pt}{20pt}
	 \textit{Step 4}& Calibrated Network-wise& $\mathcal{C}(s)_G=\frac{1}{R} \sum_{i=1}^R\log \frac{s(G)}{s\left(G_i\right)}$ & 
	 							$\mathcal{C}(c)_G=\frac{1}{R} \sum_{i=1}^R\log \frac{c(G)}{c\left(G_i\right)}$ \\  
	\bottomrule
	\end{tabular}
	\caption{The procedure of calculating the structural similarity coefficient and complementarity coefficient of a network $G$. The calibrated structural coefficients in step 4 are obtained by taking the average log ratio of network-wise coefficient over the coefficient of a sampled network $G_i$, see Eq.~\ref{eq:cali}.} \label{tab:process} 
	\end{table}

\subsubsection*{Structural coefficients}
\begin{figure}[H]
\centering
\includegraphics[width=0.7\textwidth]{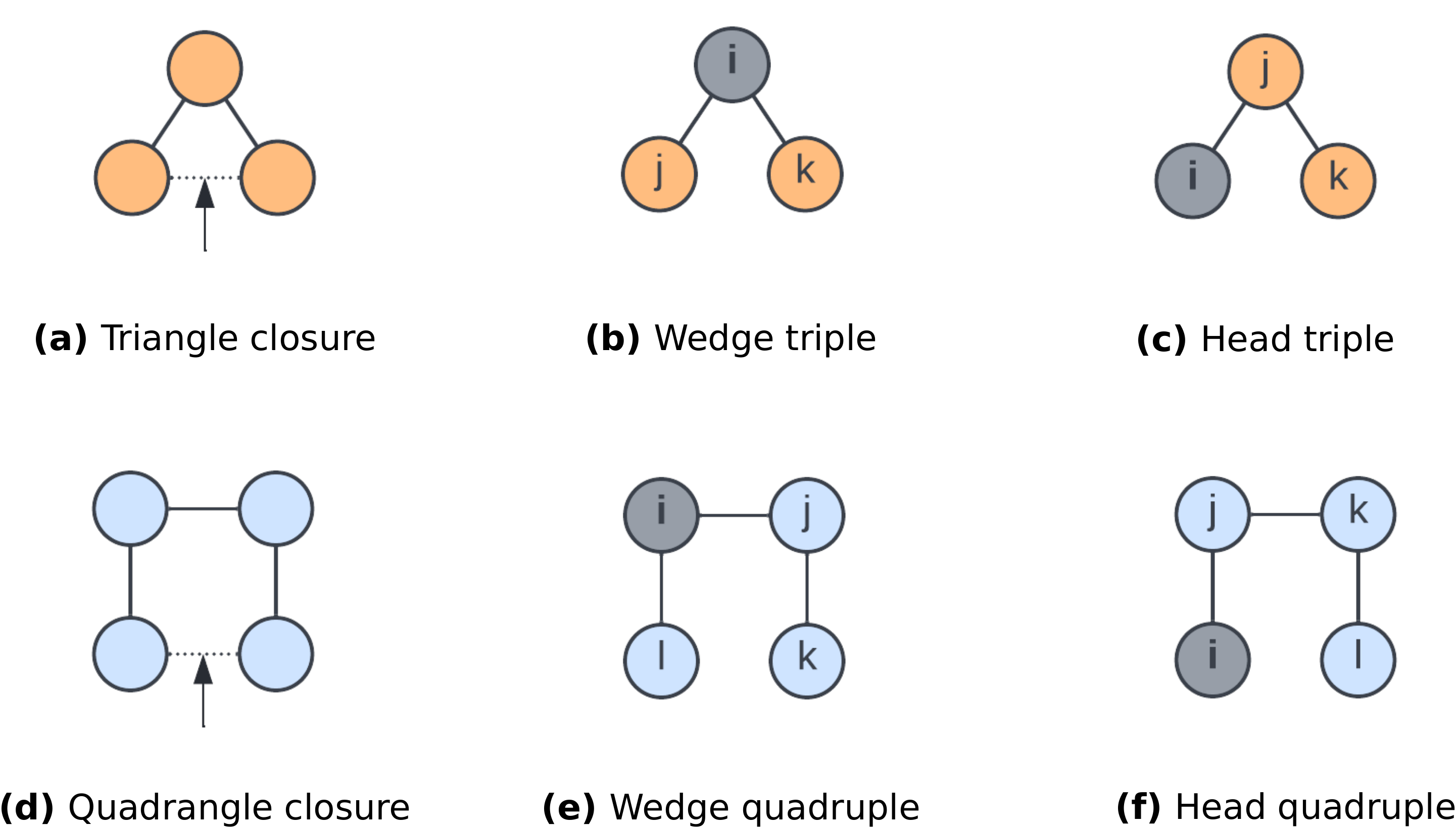}
	\caption{Quadrangle and quadruples in comparison with triangle and triples. Wedge and head triples (or quadruples) are different at where node $i$ is centered. Node $i$ in a wedge triple (b) is centered in the middle, while $i$ in a head triple (c) is centered at the beginning. Similarly, node $i$ in a wedge quadruple (e) is centered at the second location, while $i$ is at the beginning of a head quadruple (f).}
	\label{fig:motifs}
	\end{figure}

From the previous section, we learn that similarity-based networks are rich in triangles because of the triangle closure principle. 
The clustering coefficient is a classic measure of the density of triangles in a network. 
However, we cannot simply compare the number of triangles and quadrangles between two networks, because these networks have different sizes and degree distributions. 
We need to reliably calculate the statistics of triangles and quadrangles of a network to quantify similarity and complementarity. To this end, we rely on a recent work of complementarity~\cite{talaga2022}. 
The structural similarity coefficient is a weighted average of two clustering coefficients based on head and wedge triples (Figs.~\ref{fig:motifs}c and \ref{fig:motifs}b).
Analogous to the clustering coefficient, we can use structural complementarity measures based on quadrangle closure rules (Fig.~\ref{fig:motifs}d). Similarly, the structural complementarity coefficient is a weighted average of two coefficients based on head and wedge quadrangles (Figs.~\ref{fig:motifs}f and \ref{fig:motifs}e). 
The detailed procedure of calculating the structural similarity coefficient and complementarity coefficient of a network $G$ is explained in SI.

\subsection*{Calibration} \label{ap:calib}
This section presents the configuration model used to calibrate structural coefficients of semantic networks. The details of the calibration process are provided as well.

\subsubsection*{Undirected Binary Configuration Model}
In this paper, we utilize Undirected Binary Configuration Model (UBCM)~\cite{vallarano2021fast} to calibrate the structural coefficients. The UBCM generates a maximum entropy probability distribution over a network with the constraints of an expected degree sequence. It is suitable for undirected and unweighted networks. The resulting maximum entropy distributions are maximally unbiased with respect to any other property~\cite{squartini2015unbiased}. \\

\subsubsection*{Calibration of structural coefficients}\phantomsection\label{sec:cali}
First of all, we calculate one structural coefficient (similarity or complementarity) of a given network $G$. We denote this coefficient as $x(G)$. Second, we sample $R$ randomized copies $G_i$'s of the given network from the configuration model. Then, we calculate the structural coefficient $x(G_i)$ for each sampled network. At last, we take the average log ratio of $x(G)$ and $x(G_i)$'s. As a result, the calibrated coefficient $\mathcal{C}_G(x)$ based on $R$ samples from the configuration model is obtained as follows \cite{talaga2022}
	\begin{equation} \label{eq:cali}
	\mathcal{C}(x)_G=\frac{1}{R} \sum_{i=1}^R \log \frac{x(G)}{x\left(G_i\right)}.
	\end{equation} 
The calibrated structural coefficient can be less than, equal to or larger than zero. Consider the calibrated structural similarity coefficient $\mathcal{C}_G(s)$ for example:
\begin{itemize}
\item $\mathcal{C}_G(s) <0$, the structural similarity coefficient $s(G)$ is smaller than $s(G_i)$ of random networks.
\item $\mathcal{C}_G(s) =0$, the structural similarity coefficient is comparable to the ones in random networks.
\item $\mathcal{C}_G(s) >0$, the structural similarity coefficient is larger than in random networks.
\end{itemize}
	
We do not compute the structural coefficients for networks that have less than 100 nodes, because there is a high chance that there exist no triangles or quadrangles in the sampled networks and the structural coefficient $x(G_i)=0$, in that case. When $x(G_i)=0$, Eq.~\ref{eq:cali} is undefined. 

Since the runtime of the algorithm depends on the size of a network and the choice of the number of randomized networks $R$, we do not compute the structural coefficients for the two largest networks, the French `Related-To' and `Union' networks with $N>1,200,000$ each, as the computation time would be infeasible. We use $R=500$ for most networks and for the remaining two largest networks, English `Related-To' and `Union', we set $R=100$ to avoid long computation time. 

\end{document}